\documentclass{arxiv_article}
\usepackage{xspace}
\usepackage{fancyvrb}
\usepackage{array}
\usepackage{ragged2e}
\usepackage{enumitem}

\usepackage{booktabs}
\usepackage{colortbl}
\newcommand{\R}{\mathbb{R}}
\usepackage{xcolor}
\usepackage{algpseudocode}
\usepackage{adjustbox}
\usepackage{nicematrix}
\usepackage{makecell}
\usepackage{subfigure}
\usepackage{listings} 
\usepackage{float}
\usepackage{titletoc}
\usepackage{microtype}
\usepackage{graphicx}
\graphicspath{{figures/}{./}}
\usepackage{subcaption}
\usepackage{multirow}
\usepackage[table]{xcolor}
\definecolor{blue}{RGB}{0, 70, 174}
\usepackage{amsmath,amssymb,mathtools,amsthm}
\usepackage{array}
\usepackage{enumitem}

\definecolor{lightpink}{rgb}{1,0.9,0.9}
\definecolor{lightblue}{rgb}{0.8,0.9,1.0}
\definecolor{deepblue}{rgb}{0,0,0.5}
\definecolor{deepgreen}{rgb}{0,0.5,0}

\newcolumntype{C}[1]{>{\centering\arraybackslash}p{#1}}
\definecolor{oatpink}{HTML}{FDE2E4}
\definecolor{mypurple}{HTML}{4A148C}

\newcommand{\LN}{\operatorname{LN}}

\newcommand{\Sink}{\operatorname{Sinkhorn}}

\usepackage{hyperref}
\usepackage{url}
\usepackage{bm} 
\newcolumntype{Y}{>{\centering\arraybackslash}X}
\usepackage[table]{xcolor}

\usepackage{algorithm}
\usepackage{algpseudocode}
\usepackage{amsmath}
\usepackage{amssymb}
\usepackage{mathrsfs}
\usepackage{mathtools}
\usepackage{amsthm}
\usepackage{wrapfig,tabularx}

\theoremstyle{plain}
\newtheorem{theorem}{Theorem}[section]

\newtheorem{lemma}[theorem]{Lemma}

\theoremstyle{definition}

\theoremstyle{remark}
\newtheorem{remark}[theorem]{Remark}
\usepackage[textsize=tiny]{todonotes}

\renewcommand\affiliationformat[2][]{ {\footnotesize $^{#1}$#2} }
\title{HamFormer: A Hamiltonian Flow-based SE(3)-Transformer for Conformation Optimization}
\author[1]{Minjie Cheng}
\author[1]{Zhenghao Liu}
\author[1]{Yitian Wang}
\author[1]{Fanmeng Wang}
\author[2]{Anqi Dong\thanks{Corresponding authors. Email: anqidonggranet@gmail.com, hongtengxu@ruc.edu.cn}}
\author[1,3,4]{Hongteng Xu\footnotemark[1]}

\affiliation[1]{Gaoling School of Artificial Intelligence, Renmin University of China}
\affiliation[2]{Laboratory for Human-Oriented Mobility Eco-system, \'{E}cole Polytechnique F\'ed\'erale de Lausanne}
\affiliation[3]{Beijing Key Laboratory of Research on Large Models and Intelligent Governance}
\affiliation[4]{Engineering Research Center of Next-Generation Intelligent Search and Recommendation, MOE}
\begin{document}
\abstract{
The Transformer architecture plays a central role in modern deep learning, and its SE(3)-equivariant variants have been widely applied to molecular conformation modeling and optimization. Given its importance, many methods attempt to interpret and improve the Transformer through the lens of dynamics. However, they mainly focus on a single attention head rather than the full Transformer architecture and seldom consider SE(3)-equivariance in their proposed models. In this study, we propose an improved SE(3)-Transformer architecture, called HamFormer, which revisits the design of SE(3)-Transformer through the lens of Hamiltonian flow. In particular, we define a parametrized Hamiltonian flow in the phase space of position and velocity, corresponding to optimizing an energy functional that combines an optimal transport-based potential with a damped momentum. The Lie-Trotter splitting framework of the Hamiltonian flow leads to the proposed HamFormer architecture, whose feedforward computation achieves the optimization step of the proposed energy functional. Furthermore, we combine HamFormer with Equivariant Graph Neural Network (EGNN), build a new model for molecular conformation optimization. Extensive experiments on molecular ground-state conformation prediction show that HamFormer provides a strong and physically interpretable surrogate for classic SE(3)-Transformer and generally outperforms state-of-the-art competitors.
}
\maketitle
\section{Introduction}
Transformers~\citep{vaswani2017attention} have become the dominant architecture for modeling interacting elements, from tokens in language to atoms in molecules and materials. For the elements in three-dimensional space, whose underlying physics is invariant under rotations and translations, SE(3)-Transformers~\citep{thomas2018tensor,fuchs2020se3} build this symmetry into attention itself, coupling invariant scalar features with equivariant coordinates so that predictions transform consistently with their inputs. This geometric inductive bias underpins a series of practical successes across the molecular sciences, including protein structure prediction in AlphaFold~\citep{jumper2021highly,abramson2024accurate}, molecular representation learning in Uni-Mol~\citep{zhou2022uni}, equivariant force fields and materials modeling~\citep{satorras2021n,liao2023equiformer}, and ground-state conformation prediction~\citep{wang2025wgformer}. As the Transformer and its SE(3)-equivariant variants are entrusted with physically consequential predictions, understanding and improving what their layers compute has become a question of practical rather than merely theoretical significance.

Considerable efforts have gone into understanding the mechanism behind the Transformer, such as interpreting its self-attention layer through interacting particle dynamics~\citep{geshkovski2023emergence,sander2022sinkformers,wang2025wgformer} and revisiting its residual architecture as ODE flows~\citep{chen2018neural}. Building on these interpretations, some recent methods improve the Transformer architecture through the lens of dynamics. For example, Sinkformer~\citep{sander2021momentum} applies doubly stochastic attention maps based on the Sinkhorn-scaling algorithm~\citep{cuturi2013sinkhorn} and, accordingly, interprets the improved Transformer as Wasserstein gradient flows. Recent YurriFormer~\citep{zimin2026yuriiformer} and SympFormer~\citep{stein2026sympformer} introduce high-order dynamical information into the Transformer and design their models as the Nesterov-accelerated gradient descent algorithm. However, extending these dynamical interpretations to molecular refinement requires considering both feature evolution and molecular geometry. Beyond the attention operator, the architecture combines multiple attention heads, tokenwise FFNs, and pair relations that inform equivariant coordinate updates. The challenge is to connect these components within a common dynamical framework that guides their interaction while preserving SE(3)-equivariance.

In this paper, we propose \textbf{HamFormer}, which revisits the SE(3)-Transformer from the viewpoint of parametric Hamiltonian flows and improves the architecture accordingly. In principle, HamFormer is built based on coupled interacting particle dynamics, which capture the evolution of a distribution in the phase space of position and momentum. To model the dynamics, we define a Hamiltonian flow in phase space generated by an energy functional consisting of an optimal transport-based potential and a learned kinetic term. We parameterize the energy functional and discretize the resulting dynamics in the Lie--Trotter splitting framework \citep{hairer2006geometric}, with momentum retention applied before the interaction update. Given a set of particles (e.g., atomic embeddings of a molecule) and their initial momentum and pairwise relations, the discrete implementation of splitting leads to the proposed HamFormer architecture, whose feedforward computation updates the positions, momentum, and relations of the particles following the reference kick--drift structure.

Notably, HamFormer augments the atom features and pair relations used in SE(3)-equivariant molecular architectures~\citep{fuchs2020se3,zhou2022uni,wang2025wgformer} with an explicit latent momentum state, allowing interactions from earlier layers to influence subsequent feature updates. In addition, compared with existing accelerated and momentum-based Transformer architectures~\citep{zimin2026yuriiformer,stein2026sympformer,gai2026momentum}, HamFormer combines an optimal transport-based potential with multi-head feature updates and SE(3)-equivariant coordinate refinement.

\begin{figure}[t]
\centering
\includegraphics[height=3.5cm]{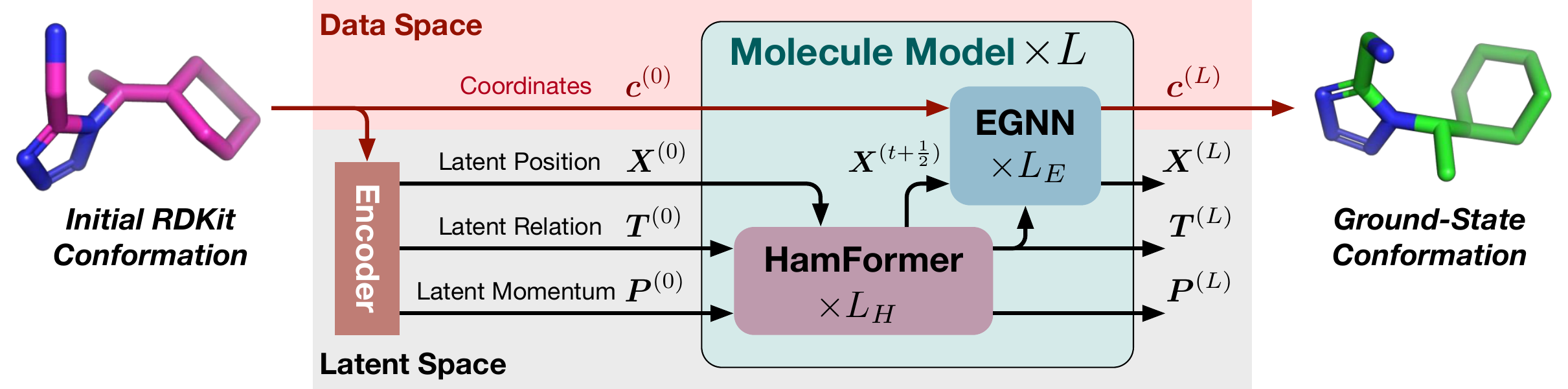}
\caption{An illustration of the proposed molecule model based on HamFormer.}
\label{fig:scheme}
\vspace{-0.14in}
\end{figure}

As shown in Figure~\ref{fig:scheme}, combining HamFormer with Equivariant Graph Neural Network (EGNN) leads to a new model for molecular conformation optimization. We evaluate the proposed model on the task of ground-state conformation prediction. Experimental results demonstrate that HamFormer provides a strong and physically interpretable surrogate for classic SE(3)-Transformer, whose feature updates follow a Hamiltonian reference structure and coordinate refinement preserves SE(3)-equivariance. Accordingly, the HamFormer-based model outperforms state-of-the-art ground-state conformation predictors in both accuracy and efficiency.
\section{Related Work and Preliminaries}
\textbf{Transformer and Its SE(3)-Equivariant Version.}
The Transformer uses attention to aggregate information across a sequence or a set of samples. Typically, given an input $\bm{X}\in\mathbb{R}^{N\times d}$, where $N$ indicates the sequence length or sample-set size and $d$ is the input feature dimension, an attention head first obtains the value, query, and key matrices by linear maps to a head dimension $d_h$, i.e., $\bm{V}=\bm{X}\bm{W}_V$, $\bm{Q}=\bm{X}\bm{W}_Q$, and $\bm{K}=\bm{X}\bm{W}_K$, with $\bm{W}_V,\bm{W}_Q,\bm{W}_K\in\mathbb{R}^{d\times d_h}$. It then aggregates $\bm{V}$ using weights determined by $\bm{Q}$ and $\bm{K}$ as $\text{Att}(\bm{V},\bm{Q},\bm{K}):=\bm{A}(\bm{Q},\bm{K})\bm{V}$, where $\bm{A}(\bm{Q},\bm{K})\in\mathbb{R}^{N\times N}$ is the attention map parametrized by $\bm{Q}$ and $\bm{K}$. The vanilla Transformer~\citep{vaswani2017attention} implements this map based on the softmax operation, i.e., $\bm{A}(\bm{Q},\bm{K})=\text{Softmax}(\bm{Q}\bm{K}^\top/\sqrt{d_h})$, where $\text{Softmax}(\cdot)$ is applied to each row, giving nonnegative attention weights that sum to one. The multi-head attention layer applies a group of linear maps, i.e., $\theta_{\mathrm{att}}=\{\bm{W}_{V,h},\bm{W}_{Q,h},\bm{W}_{K,h}\}_{h=1}^H$, to construct $H$ attention heads with a common head dimension $d_h$ and concatenates their outputs, with the output projection omitted, so that $\text{MHA}_{\theta_{\mathrm{att}}}(\bm{X}):=\text{Concat}_{\text{col}}\bigl(\{\text{Att}(\bm{V}_h,\bm{Q}_h,\bm{K}_h)\}_{h=1}^H\bigr)$, where $\text{MHA}_{\theta_{\mathrm{att}}}(\bm{X})\in\mathbb{R}^{N\times Hd_h}$, $\text{Concat}_{\text{col}}$ denotes column-wise concatenation, and $\bm{V}_h=\bm{X}\bm{W}_{V,h}$, $\bm{Q}_h=\bm{X}\bm{W}_{Q,h}$, and $\bm{K}_h=\bm{X}\bm{W}_{K,h}$ for $h=1,\ldots,H$. A complete Transformer block combines multi-head attention with a token-wise feedforward network (FFN), residual connections, and layer normalization. 

Beyond sequence modeling, attention architectures support geometric prediction. 
For molecular coordinates $\bm{C}=[\bm{c}_1,\ldots,\bm{c}_N]^\top\in\R^{N\times3}$, a coordinate predictor $f$ is SE(3)-equivariant if $f(\bm{C}\bm{R}^\top+\mathbf{1}\bm{t}^\top)=f(\bm{C})\bm{R}^\top+\mathbf{1}\bm{t}^\top$ for every rotation $\bm{R}\in\mathrm{SO}(3)$ and translation $\bm{t}\in\R^3$, with invariant atom and pair attributes held fixed. 
SE(3)-Transformer and Equiformer combine attention with equivariant representations for three-dimensional data and atomistic learning~\citep{fuchs2020se3,liao2023equiformer}.

\textbf{Improved Transformer Architectures.}
Transformer layers can be interpreted as time discretizations of differential equations or iterations of optimization algorithms. The multi-particle formulation of \citet{lu2019dynamics} associates attention and FFNs with interacting flows, while optimization-based analyses view layers as iterative updates~\citep{yang2022optimization}. Broader accounts connect Transformer dynamics to interacting-particle systems, gradient flows, and mean-field limits~\citep{geshkovski2025perspective}.

Sinkformers replace softmax attention with Sinkhorn normalization, which approximately enforces both row and column sums~\citep{sander2022sinkformers}. Entropy-regularized optimal transport provides the associated optimization problem~\citep{cuturi2013sinkhorn}. Under suitable assumptions, the resulting continuous-depth dynamics form a Wasserstein gradient flow. WGFormer applies transport-normalized attention to molecular conformation refinement and evolves pair relations across layers~\citep{wang2025wgformer}. Their gradient-flow analyses use a single attention head, although the implemented architectures can contain multiple heads. 

MomentumNets introduce persistent momentum-like states in residual networks~\citep{sander2021momentum}, and Hamiltonian neural networks learn energies governing coupled state--momentum dynamics~\citep{greydanus2019hamiltonian}. Energy Transformer derives attention and tokenwise updates from a designed energy~\citep{hoover2023energy}. More recently, YuriiFormer interprets attention and MLP updates through Lie--Trotter splitting of a composite objective~\citep{zimin2026yuriiformer}, while SympFormer derives accelerated attention from inertial dynamics on density spaces~\citep{stein2026sympformer}. Connecting these dynamical descriptions to molecular refinement requires accounting for both invariant feature processing and equivariant coordinate updates.
\section{Proposed HamFormer Architecture}
{In HamFormer, a parametrized Hamiltonian reference flow couples invariant atom features with latent momentum.} The potential force updates momentum, and the kinetic gradient advances the features. {Its Lie--Trotter splitting determines the feedforward update structure for molecular conformation optimization.} Evolving multi-head relations condition these interactions and provide invariant edge information to EGNN for equivariant coordinate refinement.

\subsection{Hamiltonian Flow}
Hamiltonian dynamics are specified by an energy and a canonical state--momentum pairing~\citep[Sec.~3.3.1]{evans2022partial}. On a periodic domain $\Omega$ of dimension $d$, let $\bm{x}\in\Omega$ denote position, $\rho$ a smooth probability density, and $S$ its conjugate potential, with velocity field $\nabla S$. Let $\mathcal F$ be a potential-energy functional with first variation $\mathcal F'$. For $N$ particles, collect positions $\bm{x}_i$ and momentum $\bm{p}_i$ as rows of $\bm{X},\bm{P}\in\R^{N\times d}$, and let $\rho_N:=N^{-1}\sum_{i=1}^N\delta_{\bm{x}_i}$ be the empirical approximation of $\rho$, where $\delta_{\bm{x}}$ denotes the Dirac measure at $\bm{x}$. The Wasserstein Hamiltonian~\citep{chow2020wasserstein} and its particle counterpart are
\begin{equation}
\mathcal H_W(\rho,S)\!=\!\frac12\int_\Omega\|\nabla S(\bm{x})\|^2\rho(\bm{x})\,\mathrm d\bm{x}\!+\!\mathcal F(\rho)\;\mbox{ and }\;H_N(\bm{X},\bm{P})\!=\!\frac12\sideset{}{_{i=1}^N}\sum\|\bm{p}_i\|^2\!+\!N\mathcal F(\bm{X}),
\label{eq:hf-density-energy}
\end{equation}
where the two terms of $H_N$ define the kinetic energy $K_N(\bm{P})$ and potential energy $V_N(\bm{X})$, respectively. The particle expression uses total-energy normalization and assumes $\mathcal F(\bm{X})$ is well defined and smooth in $\bm{X}$.

Writing $\rho_t,S_t$ for the state at time $t$, define the momentum along a characteristic trajectory $\bm{x}(t)$ by $\bm{p}(t)=\nabla S_t(\bm{x}(t))$ and the phase-space Hamiltonian by $h_t(\bm{x},\bm{p})=\tfrac12\|\bm{p}\|^2+\mathcal F'(\rho_t)(\bm{x})$. With the time-dependent additive constant in $S_t$ chosen appropriately, the density and potential satisfy the continuity and Hamilton--Jacobi equations, whose characteristic trajectories obey Hamilton's equations, i.e.,
\begin{equation*}
\resizebox{\linewidth}{!}{$\displaystyle \begin{cases}\partial_t\rho_t+\nabla\cdot(\rho_t\nabla S_t)=0&\text{(continuity)}\\[2pt]\partial_tS_t+\tfrac12\|\nabla S_t\|^2+\mathcal F'(\rho_t)=0&\text{(HJE)}\end{cases}\begin{cases}\dot{\bm{x}}=\nabla_{\bm{p}}h_t=\bm{p}\\[2pt]\dot{\bm{p}}=-\nabla_{\bm{x}}h_t=-\nabla\mathcal F'(\rho_t)(\bm{x})\end{cases}\hspace{-0.1in}\text{(Hamilton)},$}
\end{equation*}
where the right-hand system is evaluated along $(\bm{x}(t),\bm{p}(t))$, with probability mass preserved for smooth periodic solutions and $\mathcal H_W$ conserved when $\mathcal F$ is time-independent (see also, Appendix~\ref{app:hamiltonian-dynamics} for more details.).

More generally, for smooth, time-independent kinetic and potential energies $K_N,V_N$, the potential subflow changes momentum at fixed positions, while the kinetic subflow advances positions at fixed momentum. Their Lie--Trotter composition gives first-order \emph{symplectic Euler}~\citep[see also,][Sec.~4]{wu2025parameterized}. For a step size $\eta>0$, let $\bm{X}^k,\bm{P}^k$ approximate the states at $t_k=k\eta$. The particle dynamics and their discretization are
\begin{equation}
\begin{cases}\dot{\bm{P}}=-\nabla_{\bm{X}}V_N(\bm{X})\\[2pt]\dot{\bm{X}}=\nabla_{\bm{P}}K_N(\bm{P})\end{cases}\ \text{(continuous)}\quad\begin{cases}\bm{P}^{k+1}=\bm{P}^k-\eta\nabla_{\bm{X}}V_N(\bm{X}^k)\\[2pt]\bm{X}^{k+1}=\bm{X}^k+\eta\nabla_{\bm{P}}K_N(\bm{P}^{k+1})\end{cases}\text{(discretized)}
\label{eq:hf-symplectic-euler}
\end{equation}
with the second-order \emph{St\"ormer--Verlet} scheme obtained by symmetrizing this kick--drift splitting~\citep{hairer2006geometric}.

\subsection{Generalizing SE(3)-Transformer to Parametrized Hamiltonian Flows}
\textbf{A Parametric Instantiation of Hamiltonian.}
With $\theta$ denoting the model parameters, we retain the canonical structure of~(\ref{eq:hf-density-energy}), replacing the quadratic kinetic energy by a learned density $\kappa_\theta$ and the potential by a transport value. The rows of $\bm{X},\bm{P}$ now represent invariant atom features and latent momentum. Let $\mu$ be a probability measure over $(\xi,\bm{x},\bm{p})\in[0,1]\times\R^d\times\R^d$ with finite second moments, and let $\nu$ be its marginal on $(\xi,\bm{x})$. The fixed auxiliary label $\xi$ preserves pair identity as features evolve and is used only in this distributional representation. For each head $h$, a bounded nonnegative measurable kernel $t_h(\xi,\xi')$ specifies the relations, with $t=(t_h)_{h=1}^H$ fixed during the reference dynamics.

The potential is the optimal value of a coupling problem, whose optimizer supplies the interaction weights. Let $\bm{z}_h$ project features to head dimension $d_h$. With $\Pi(\nu,\nu)$ denoting couplings with both marginals $\nu$ and $\operatorname{KL}$ denoting relative entropy, define
\begin{align}
a_{\theta,h}(\xi,\bm{x},\xi',\bm{x}'\mid t_h)&=\bigl[\bm{z}_h(\bm{x})^\top\bm{z}_h(\bm{x}')\bigr]/\sqrt{d_h}-\log\bigl(t_h(\xi,\xi')+\varepsilon_{\mathrm{num}}\bigr),\notag\\ \mathcal V_{\varepsilon_{\mathrm{OT}},h}[\nu\mid t_h]&=\sideset{}{_{\pi_h\in\Pi(\nu,\nu)}}\sup\bigl\{\int a_{\theta,h}\,\mathrm d\pi_h-\varepsilon_{\mathrm{OT}}\operatorname{KL}(\pi_h\|\nu\otimes\nu)\bigr\},
\label{eq:mean-field-interaction}
\end{align}
where $\bm{z}_h(\bm{x})=\bm{W}_h^\top\bm{x}+\bm{b}_h$ for $\bm{W}_h\in\R^{d\times d_h}$ and $\bm{b}_h\in\R^{d_h}$. The constants $\varepsilon_{\mathrm{num}}>0$ and $\varepsilon_{\mathrm{OT}}>0$ stabilize the logarithm and control entropic regularization, respectively. This is entropy-regularized optimal transport with cost $-a_{\theta,h}$, see also~\citep{cuturi2013sinkhorn}.

Extending the empirical representation above, let $\mu_N=N^{-1}\sum_{i=1}^N\delta_{(\xi_i,\bm{x}_i,\bm{p}_i)}$ for distinct labels $\xi_i$, with labeled feature marginal $\nu_N$. Set $(\bm{T}_h)_{ij}=t_h(\xi_i,\xi_j)$ and $\bm{\mathsf{T}}=(\bm{T}_h)_{h=1}^H$. Couplings in $\Pi(\nu_N,\nu_N)$ have masses $\pi_{ij}/N$, where the transport matrix $\bm{\pi}=(\pi_{ij})$ belongs to $\bm{\Pi}:=\{\bm{\pi}\in\R_+^{N\times N}:\bm{\pi}\mathbf{1}=\bm{\pi}^\top\mathbf{1}=\mathbf{1}\}$. For a score matrix $\bm{S}\in\R^{N\times N}$, the finite transport value is
\begin{equation}
\Phi_{\varepsilon_{\mathrm{OT}}}(\bm{S})=\sideset{}{_{\bm{\pi}\in\bm{\Pi}}}\max\sideset{}{_{i,j}}\sum\pi_{ij}\bigl[S_{ij}-\varepsilon_{\mathrm{OT}}(\log\pi_{ij}-1)\bigr].
\label{eq:ot-value}
\end{equation}
Writing $\bm{Z}_h$ for the projected feature matrix, the distributional and particle Hamiltonians are
\begin{align}
\mathcal H_\theta[\mu| t]&\!=\!\int\kappa_\theta(\bm{p})\,\mathrm d\mu+\frac12\sideset{}{_{h=1}^H}\sum\sqrt{d_h}\mathcal V_{\varepsilon_{\mathrm{OT}},h}[\nu\mid t_h]=:\mathcal K_\theta[\mu]+\mathcal V_\theta[\nu\mid t],\label{eq:mean-field-hamiltonian}\\ 
H_{\theta,N}(\bm{X},\bm{P}|\bm{\mathsf{T}})&\!=\!\sideset{}{_{i=1}^N}\sum\kappa_\theta(\bm{p}_i)+\frac12\sideset{}{_{h=1}^H}\sum\sqrt{d_h}\Phi_{\varepsilon_{\mathrm{OT}}}([\bm{Z}_h\bm{Z}_h^\top]/\sqrt{d_h}-\log(\bm{T}_h+\varepsilon_{\mathrm{num}})),
\label{eq:param-hamiltonian}
\end{align}
where $\mathcal K_\theta,\mathcal V_\theta$ denote the kinetic and potential energy functionals, and the two terms of the particle Hamiltonian define $K_\theta(\bm{P})$ and $V_\theta(\bm{X}\mid\bm{\mathsf{T}})$. The particle Hamiltonian equals $N\mathcal H_\theta[\mu_N\mid t]$ up to a constant independent of $\bm{X},\bm{P}$, as in the total-energy normalization of~(\ref{eq:hf-density-energy}), with the exact relation derived in Appendix~\ref{app:particle-representation}. The projected features are $\bm{Z}_h=\bm{X}\bm{W}_h+\mathbf{1}\bm{b}_h^\top$, with matrix logarithms and scalar offsets applied entrywise. 
{The kinetic density is ${\kappa_\theta(\bm p)=\tfrac12\|\sigma_{\sqrt{0.1}}(\bm B^\top\bm p+\bm b)\|_2^2}$ for learned $\bm B\in\R^{d\times d}$ and $\bm b\in\R^d$, {where $\sigma_a$ denotes entrywise leaky-ReLU with negative slope $a$. The slope $\sqrt{0.1}$ ensures that $\nabla_{\bm p}\kappa_\theta(\bm p)=\bm B\,\sigma_{0.1}(\bm B^\top\bm p+\bm b)$, matching the kinetic drift.}}
We take $\phi_{0.1}(u)=\int_0^u\sigma_{0.1}(s)\,\mathrm ds$, where $\sigma_{0.1}(u)=\max\{u,0.1u\}$ is leaky-ReLU, applied entrywise.

\textbf{Lie--Trotter realization.}
Fix $\theta$ and $\bm{\mathsf{T}}$, and let $\bm{\pi}_h^\star$ be the unique optimizer of~(\ref{eq:ot-value}) for head $h$. The envelope identity $\nabla_{\bm{S}}\Phi_{\varepsilon_{\mathrm{OT}}}(\bm{S})=\bm{\pi}^\star$, obtained from Danskin's theorem~\citep{danskin1966theory}, together with $\phi_{0.1}'=\sigma_{0.1}$, gives
\begin{equation*}
\nabla_{\bm{P}}K_\theta(\bm{P})=\sigma_{0.1}(\bm{P}\bm{B}+\mathbf{1}\bm{b}^\top)\bm{B}^\top\quad\text{and}\quad\nabla_{\bm{X}}V_\theta(\bm{X}\mid\bm{\mathsf{T}})=\frac12\sideset{}{_{h=1}^H}\sum\bigl(\bm{\pi}_h^\star+(\bm{\pi}_h^\star)^\top\bigr)\bm{Z}_h\bm{W}_h^\top,
\end{equation*}
where the prefactor $\sqrt{d_h}/2$ in the potential accounts for score normalization and the two feature arguments. Symmetric relations imply symmetric optimal plans, reducing the force to $-\sum_h\bm{\pi}_h^\star\bm{Z}_h\bm{W}_h^\top$. Substitution into~(\ref{eq:hf-symplectic-euler}) gives the reference kick--drift step, with momentum retention incorporated by applying $\bm{P}\leftarrow e^{-\gamma\eta}\bm{P}$ before the kick, where $\gamma\geq0$ and $\gamma=0$ recovers the conservative splitting. Appendices~\ref{app:reference-gradients} and~\ref{app:splitting} derive the gradients and splitting, respectively.

\begin{wrapfigure}{r}{0.45\textwidth}
\centering
\includegraphics[width=\linewidth]{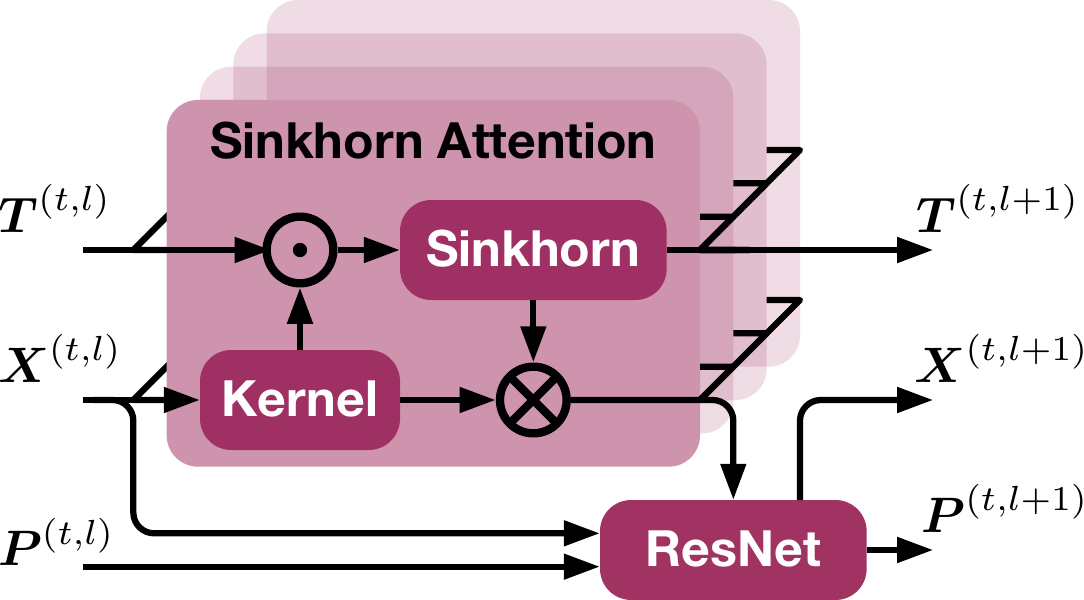}
\caption{HamFormer architecture.}
\label{fig:hamformer}
\vspace{-0.23in}
\end{wrapfigure}
\textbf{Implementation.}
HamFormer uses finite Sinkhorn iterations for transport, normalized learned maps for the kick and drift, and a tokenwise residual feed-forward network (FFN). Its stages alternate with EGNN stages, which use the final relations as invariant edge information for SE(3)-equivariant coordinate refinement, as shown in Figure~\ref{fig:hamformer}.

Let $\bm{c}_i^{\mathrm{in}}\in\R^3$ be the input coordinates and $D_{ij}^{\mathrm{in}}=\|\bm{c}_i^{\mathrm{in}}-\bm{c}_j^{\mathrm{in}}\|$. For each head, let $g_h(e_{ij})$ embed the unordered atom-pair type $e_{ij}$ into a scalar. Write $\Sink_{\varepsilon_{\mathrm{OT}},I}(\bm{S})$ for $I$ alternating row and column scaling iterations on $\exp(\bm{S}/\varepsilon_{\mathrm{OT}})$, targeting marginals $\mathbf{1}/N$. Assuming $\max_{u,v}D_{uv}^{\mathrm{in}}>0$, initialize
\begin{equation}
\scalebox{0.97}{$\displaystyle \bm{T}_h^{(0)}=N\,\Sink_{\varepsilon_{\mathrm{OT}},I}\Bigl(\frac{[g_h(e_{ij})D_{ij}^{\mathrm{in}}]_{ij}}{\max_{u,v}D_{uv}^{\mathrm{in}}}\Bigr)$}
\label{eq:initial-relation}
\end{equation}
that $N$ rescales the target row and column sums to one, and we use $I=3$ iterations to approximate these constraints. At layer $\ell$, use layer-indexed parameters and let $\LN_{\mathrm{in},\ell}$, $\LN_{P,\ell}$, and $\LN_{X,\ell}$ denote LayerNorm on the input features, interaction message, and drift. The relation update is
\begin{equation*}
\begin{aligned}
\bm{Z}_h^{(\ell)}&=\LN_{\mathrm{in},\ell}(\bm{X}^{(\ell)})\bm{W}_h^{(\ell)}+\mathbf{1}(\bm{b}_h^{(\ell)})^\top,\\ \bm{T}_h^{(\ell+1)}&=N\,\Sink_{\varepsilon_{\mathrm{OT}},I}\Bigl(\bm{Z}_h^{(\ell)}(\bm{Z}_h^{(\ell)})^\top/\sqrt{d_h}-\log(\bm{T}_h^{(\ell)}+\varepsilon_{\mathrm{num}})\Bigr),\end{aligned}
\end{equation*}
with the updated relations supplying the current interaction weights and conditioning the next layer. At $\varepsilon_{\mathrm{OT}}=1$, the kernel divides the exponentiated affinity entrywise by $\bm{T}_h^{(\ell)}+\varepsilon_{\mathrm{num}}$. The initialization and updates depend only on invariant attributes, distances, and features, so the relations remain invariant under rigid motions.

With learned retention $\alpha_\ell\in\R$ and post-drift features $\bm{U}$, the state update is
\begin{equation}
\bm{P}^{(\ell+1)}\!=\!-\!\tanh(\alpha_\ell)\bm{P}^{(\ell)}\!+\!\LN_{P,\ell}\Bigl(-\!\sideset{}{_{h=1}^H}\sum\bm{T}_h^{(\ell+1)}\bm{Z}_h^{(\ell)}(\bm{W}_h^{(\ell)})^\top\Bigr),~~\bm{X}^{(\ell+1)}\!=\!\bm{U}\!+\!\operatorname{FFN}_\ell(\bm{U}),
\label{eq:implemented-kick}
\end{equation}
where $\bm{U}=\bm{X}^{(\ell)}+\LN_{X,\ell}\bigl[\sigma_{0.1}(\bm{P}^{(\ell+1)}\bm{B}_\ell+\mathbf{1}\bm{b}_\ell^\top)\bm{B}_\ell^\top\bigr]$. Only the interaction branch of the momentum update is normalized. The coefficient $-\tanh(\alpha_\ell)\in(-1,1)$ is shared within each layer and allows signed retention, with $\alpha_\ell=0$ at initialization. Also, it has no effect in the first layer because the initial momentum is $\bm{P}^{(0)}=\bm{0}$. Learned maps and normalizations replace explicit step sizes.

For hidden width $d_{\mathrm{FF}}$, let $\bm{W}_{1,\ell}\in\R^{d\times d_{\mathrm{FF}}}$ and $\bm{W}_{2,\ell}\in\R^{d_{\mathrm{FF}}\times d}$ have compatible biases $\bm{b}_{1,\ell},\bm{b}_{2,\ell}$. With input LayerNorm $\LN_{F,\ell}$, activation and output dropout $\mathrm{Drop}_{a,\ell},\mathrm{Drop}_{o,\ell}$, and hidden activation $\bm{G}$, define $\operatorname{FFN}_\ell(\bm{U})=\mathrm{Drop}_{o,\ell}\bigl(\bm{G}\bm{W}_{2,\ell}+\mathbf{1}\bm{b}_{2,\ell}^\top\bigr)$, where $\bm{G}=\mathrm{Drop}_{a,\ell}\bigl[\operatorname{GELU}(\LN_{F,\ell}(\bm{U})\bm{W}_{1,\ell}+\mathbf{1}\bm{b}_{1,\ell}^\top)\bigr]$. 

Before output normalization, the tied drift equals the reference kinetic gradient. With fixed relations and exact symmetric transport plans, the interaction message equals the reference force with respect to normalized features. 
Training differentiates through the relation updates, normalization, and Sinkhorn iterations using the supervised loss (see also, Appendix \ref{app:conformation-optimization} for more details).

{The combined gradient of the multi-head OT potential plays the interaction role of multi-head attention in an SE(3)-Transformer: it supplies the force that updates latent momentum. The tied tokenwise feedforward map realizes the kinetic gradient and advances features using the updated momentum. Together, these operations form the reference kick--drift structure. For fixed parameters and relations, quadratic kinetic energy gives $\ddot{\bm X}=-\nabla_{\bm X}V_\theta$, introducing second-order feature dynamics in contrast to the first-order flows in the single-head analyses of Sinkformers and WGFormer~\citep{sander2022sinkformers,wang2025wgformer}. The learned kinetic energy generalizes the momentum-to-velocity relation. The residual FFN and EGNN extend this reference through learned feature corrections and equivariant coordinate updates. This provides a physics-aware organization of the complete architecture, connecting attention, momentum, feedforward processing, and molecular geometry.}

\textbf{Equivariant coordinate refinement.}
The initial relations depend only on atom-pair types and distances, and HamFormer preserves invariance of the features, momentum, and relations under rigid motions. Each EGNN stage~\citep{satorras2021n} uses these invariant quantities to construct coordinate displacements. At EGNN layer $q$, learned pair messages produce a residual feature update using atom features, current and stage-entry squared distances, and final HamFormer relations. A separate MLP uses the updated features and the same distances and relations to produce invariant scalar weights $w_{ij}^{(q)}$. The stage-entry distances and relations remain fixed throughout the stage. Coordinates are updated by $\bm{c}_i^{(q+1)}=\bm{c}_i^{(q)}+\sum_j w_{ij}^{(q)}(\bm{c}_i^{(q)}-\bm{c}_j^{(q)})/(\|\bm{c}_i^{(q)}-\bm{c}_j^{(q)}\|+\varepsilon_{\mathrm{dist}})$, where $\varepsilon_{\mathrm{dist}}>0$ stabilizes the denominator and scalar normalization factors are absorbed into $w_{ij}^{(q)}$. The scalar coefficients are invariant, while relative vectors rotate and are unchanged by translation, giving an SE(3)-equivariant update. Appendix~\ref{app:equivariance} proves equivariance of the complete pipeline. EGNN leaves momentum unchanged and passes its feature output directly to the next HamFormer stage. Figure~\ref{fig:scheme} shows the complete architecture.

\begin{wraptable}{r}{0.62\textwidth}
\vspace{-0.17in}
\centering
\small
\setlength{\abovecaptionskip}{2pt}
\setlength{\belowcaptionskip}{2pt}
\caption{{Physical roles of HamFormer components.}}
\label{tab:physical-correspondence}
\footnotesize
\setlength{\tabcolsep}{2pt}
\renewcommand{\arraystretch}{1.1}
\resizebox{\linewidth}{!}{%
\begin{tabular}{@{}llccl@{}}
\toprule
Component & {Form} & {Physics} & {State} & {Molecular Role} \\
\midrule
{OT Attention} & {$-\nabla_{\bm X}V_\theta$} & {Force} & {$\bm P$} & {Atom Interactions} \\
{Kinetic Map} & {$\nabla_{\bm P}K_\theta$} & {Velocity} & {$\bm X$} & {Feature Motion} \\
Retention & {$-\tanh(\alpha_\ell)\bm P$} & {Signed Carry} & {$\bm P$} & {Refinement Memory} \\
Relations & {$\bm T_h^{(\ell)}\mapsto\bm T_h^{(\ell+1)}$} & {Coupling} & {$\bm{\mathsf T}$} & {Pair Information} \\
EGNN & {$\bm c_i-\bm c_j$} & {Direction} & {$\bm C$} & {Coordinate Refinement} \\
\bottomrule
\end{tabular}%
\vspace{-0.18in}
}
\end{wraptable}

\noindent\leavevmode\textbf{Physical interpretation.}
{Compared with the first-order transport-flow interpretations of Sinkformers and WGFormer~\citep{sander2022sinkformers,wang2025wgformer}, HamFormer introduces an independent momentum state between attention and feature motion. The combined gradient of the multi-head transport potential supplies the momentum kick, while the tied feedforward map supplies the kinetic drift. For fixed parameters and relations, quadratic kinetic energy gives $\ddot{\bm X}=-\nabla_{\bm X}V_\theta$, so the interaction force determines acceleration. The learned kinetic energy generalizes the momentum-to-velocity relation. For ground-state conformation prediction, momentum carries earlier interaction effects across refinement steps, while evolving relations retain atom-pair information. Transport couples the atom features across the molecule, and EGNN uses the resulting features and relations to produce equivariant coordinate displacements~\citep{satorras2021n}. Table~\ref{tab:physical-correspondence} summarizes these roles. The mechanical interpretation concerns latent features and momentum, with molecular coordinates updated by EGNN.}

\begin{table}[h]
\centering
\small
\setlength{\tabcolsep}{3pt}
\renewcommand{\arraystretch}{1.25}
\caption{{Reference dynamics under the respective theoretical assumptions and features of the reported architectures. OT denotes entropic optimal transport, and pair memory denotes relations retained across layers. The SympFormer row reports its softmax variant.}}
\label{tab:model-comparison}
\resizebox{\linewidth}{!}{%
\begin{tabular}{@{}llcccc@{}}
\toprule
Model & Reference dynamics & Interaction & \shortstack{Momentum\\parametrization} & \shortstack{Pair\\memory} & SE(3) \\
\midrule
Sinkformers~\citeyearpar{sander2022sinkformers}
& Wasserstein PDE: $\partial_t\rho=\nabla\cdot(\rho\nabla\mathcal F'(\rho))$
& OT & -- & $\times$ & $\times$ \\
\addlinespace
WGFormer~\citeyearpar{wang2025wgformer}
& Wasserstein PDE: $\partial_t\rho=\nabla\cdot(\rho\nabla\mathcal F'(\rho))$
& OT & -- & $\checkmark$ & $\checkmark$ \\
\addlinespace
YuriiFormer~\citeyearpar{zimin2026yuriiformer}
& Nesterov: $\Delta\bm{X}^{k+1}=\beta_k\Delta\bm{X}^{k}-\eta_k\nabla E(\bm{X}^{k}+\alpha_k\Delta\bm{X}^{k})$
& Softmax & Learned coefficients & $\times$ & $\times$ \\
\addlinespace
SympFormer~\citeyearpar{stein2026sympformer}
& Hamiltonian PDE: $\partial_t\rho=\mathcal H'_S,\quad \partial_tS=-\mathcal H'_\rho-\gamma(t)S$
& Softmax & Metric, damping & $\times$ & $\times$ \\
\addlinespace
HamFormer \textbf{(ours)}
& Hamiltonian PDE: $\partial_t\mu+\{\mu,h_\mu\}=0$
& OT & Kinetic map, signed retention & $\checkmark$ & $\checkmark$ \\
\bottomrule
\end{tabular}%
}
\end{table}

{Table~\ref{tab:model-comparison} compares the reference dynamics and architectural features. YuriiFormer introduces momentum through a Nesterov template, while SympFormer uses an attention-dependent kinetic metric. HamFormer combines a separable reference Hamiltonian at fixed relations with entropic transport interactions, persistent pair relations, and SE(3)-equivariant coordinate refinement.}

{In the Nesterov template, $\Delta\bm X^k=\bm X^k-\bm X^{k-1}$, and $\alpha_k,\beta_k,\eta_k$ control lookahead, retention, and step size for the reference objective $E$. In the Hamiltonian rows, primes denote first variations, and $h_\mu=\mathcal H'_\theta[\mu\mid t]$ is evaluated at fixed parameters and relations. The canonical Poisson bracket $\{f,g\}=\nabla_{\bm x}f\cdot\nabla_{\bm p}g-\nabla_{\bm p}f\cdot\nabla_{\bm x}g$ describes transport in feature--momentum space.}

{With feature dimensions, head count, and Sinkhorn iterations fixed, each dense HamFormer layer requires $O(N^2)$ time and storage, as detailed in Remark~\ref{rem:complexity}. 
In practice, we use three Sinkhorn iterations and reuse each head's projected features for both score construction and transport aggregation.}
The practical layer retains the reference kick--drift order, with finite Sinkhorn iterations, normalized learned maps, signed retention, and a residual FFN.
Appendix~\ref{app:conformation-optimization} gives the complete pipeline, both algorithms, and the supervised training objective.

\section{Experiments}
To validate the effectiveness and physical interpretability of HamFormer, we evaluate our model on molecular ground-state conformation prediction following the established experimental protocol of WGFormer~\cite{wang2025wgformer}. 
In addition to quantitative comparisons with existing 2D graph models and 3D equivariant architectures, we provide an empirical analysis of the latent dynamics learned by HamFormer, focusing on the layerwise evolution of phase-space energy and the role of the latent momentum state.

\subsection{Experimental Setup}
\textbf{Datasets and splitting strategy.}
We evaluate our model on two standard molecular conformation benchmarks: QM9 and the larger-scale Molecule3D~\citep{xu2021molecule3d}. QM9 comprises approximately 134k small organic molecules with up to nine heavy atoms, with ground-state geometries calculated via density functional theory (DFT). Molecule3D contains substantially larger and more flexible drug-like molecules. Following the benchmark protocol in \citet{wang2025wgformer}, we adopt the standard training, validation, and test splits for QM9, and evaluate on the standard random split for Molecule3D.

\textbf{Baselines.}
Following \citet{wang2025wgformer}, we compare HamFormer with two categories of baseline methods: 
(1) 2D molecular graph models, including GINE~\citep{hu2020pretraining}, GATv2~\citep{brody2021attentive}, GPS~\citep{rampavsek2022recipe}, and GTMGC~\citep{xu2023gtmgc}, which generate 3D coordinates directly from 2D molecular graphs; and 
(2) 3D conformation refinement methods, including ConfOpt with TwoAtom and ThreeAtom energy functions~\citep{guan2022energy}, ConfOpt variants using SE(3)-Transformer~\citep{fuchs2020se3} and EGNN~\citep{satorras2021n} backbones, and WGFormer~\citep{wang2025wgformer}.

\textbf{Evaluation metrics.}
Following the established evaluation protocol in \citet{xu2021molecule3d,wang2025wgformer}, we evaluate predicted conformations using three complementary geometric metrics: 
(1) \textbf{D-MAE} (Distance Mean Absolute Error), (2) \textbf{D-RMSE} (Distance Root Mean Square Error), which measure the accuracy of internal molecular geometry via pairwise interatomic distances, and 
(3) \textbf{C-RMSD} (Coordinate Root Mean Square Deviation), which assesses global 3D structural alignment by computing the coordinate error against the ground truth after optimal rotation and translation. 
All metrics are reported in \AA{}ngstr\"oms (\AA{}), where lower values indicate better predictive quality.

\subsection{Main Results}
Table~\ref{tab:main_results} reports the quantitative comparison on Molecule3D and QM9. 
Compared to WGFormer~\citep{wang2025wgformer}, HamFormer achieves lower errors across all metrics on QM9 (e.g., test C-RMSD of 0.198\,\AA{} vs. 0.206\,\AA{}), while delivering improved D-MAE and C-RMSD on Molecule3D. 
HamFormer attains these results with 16.3M parameters, an approximate 79\% reduction compared to WGFormer (78.9M). 
Compared to the first-order Wasserstein gradient flow in WGFormer, incorporating second-order phase-space dynamics provides a more parameter-efficient inductive bias for molecular conformation refinement.

\begin{table}[htbp]
\centering
\caption{Performance comparison on Molecule3D and QM9 datasets. 
$\downarrow$ indicates that lower values are better. 
\textbf{Bold} indicates the best performance, and \underline{underline} indicates the second best.
Baseline results follow~\cite{wang2025wgformer}.
}
\label{tab:main_results}
\setlength{\tabcolsep}{3pt}
\resizebox{\linewidth}{!}{
\begin{tabular}{cc c ccc ccc}
\toprule
\multirow{2}{*}{Dataset} 

& \multirow{2}{*}{Model}
& \multicolumn{3}{c}{Validation}
& \multicolumn{3}{c}{Test} 
\\
\cmidrule(lr){3-5}
\cmidrule(lr){6-8}
&
& D-MAE $\downarrow$ & D-RMSE $\downarrow$ & C-RMSD $\downarrow$
& D-MAE $\downarrow$ & D-RMSE $\downarrow$ & C-RMSD $\downarrow$ 
\\
\midrule
\multirow{10}{*}{\shortstack{Molecule3D
\\(Random)}}
& GINE(2D)
& 0.590 & 1.014 & 1.116
& 0.592 & 1.018 & 1.116 \\
& GATv2(2D)
& 0.563 & 0.983 & 1.082
& 0.564 & 0.986 & 1.083 \\
& GPS(2D)
& 0.528 & 0.909 & 1.036
& 0.529 & 0.911 & 1.038 \\
& GTMGC(2D)
& 0.432 & 0.719 & 0.712
& 0.433 & 0.721 & 0.713 \\
& SE(3)-Transformer
& 0.466 & 0.712 & 0.800
& 0.467 & 0.774 & 0.802 \\
& EGNN
& 0.461 & 0.704 & 0.798
& 0.462 & 0.766 & 0.799 \\
& ConfOpt-TwoAtom
& 0.438 & 0.668 & 0.748
& 0.438 & 0.670 & 0.749 \\
& ConfOpt-ThreeAtom
& 0.429 & 0.659 & 0.734
& 0.430 & 0.661 & 0.736 \\
& WGFormer 
& \underline{0.391}
& \textbf{0.649} 
& \underline{0.662}
& \underline{0.392}
& \textbf{0.652} 
& \underline{0.664}
\\
& \cellcolor[RGB]{226,230,235}HamFormer
& \cellcolor[RGB]{226,230,235}\textbf{0.386}
& \cellcolor[RGB]{226,230,235}\underline{0.653}
& \cellcolor[RGB]{226,230,235}\textbf{0.652}
& \cellcolor[RGB]{226,230,235}\textbf{0.387}
& \cellcolor[RGB]{226,230,235}\underline{0.656}
& \cellcolor[RGB]{226,230,235}\textbf{0.654}
\\
\midrule
\multirow{10}{*}{QM9}
& GINE(2D)
& 0.357 & 0.673 & 0.685
& 0.357 & 0.669 & 0.693 \\
& GATv2(2D)
& 0.339 & 0.663 & 0.661
& 0.339 & 0.659 & 0.666 \\
& GPS(2D)
& 0.326 & 0.644 & 0.662
& 0.326 & 0.640 & 0.666 \\
& GTMGC(2D)
& 0.262 & 0.468 & 0.362
& 0.264 & 0.470 & 0.367 \\
& SE(3)-Transformer
& 0.254 & 0.451 & 0.296
& 0.256 & 0.455 & 0.303 \\
& EGNN
& 0.248 & 0.442 & 0.257
& 0.251 & 0.449 & 0.265 \\
& ConfOpt-TwoAtom
& 0.245 & 0.439 & 0.245
& 0.248 & 0.444 & 0.254 \\
& ConfOpt-ThreeAtom
& 0.241 & 0.433 & 0.237
& 0.244 & 0.438 & 0.246 \\
& WGFormer
& \underline{0.223} 
& \underline{0.416}
& \underline{0.198}
& \underline{0.227}
& \underline{0.422}
& \underline{0.206}
\\
& \cellcolor[RGB]{226,230,235}HamFormer
& \cellcolor[RGB]{226,230,235}\textbf{0.215}
& \cellcolor[RGB]{226,230,235}\textbf{0.414}
& \cellcolor[RGB]{226,230,235}\textbf{0.191}
& \cellcolor[RGB]{226,230,235}\textbf{0.218}
& \cellcolor[RGB]{226,230,235}\textbf{0.419}
& \cellcolor[RGB]{226,230,235}\textbf{0.198}
\\
\bottomrule
\end{tabular}
}
\end{table}

\subsection{Ablation Studies and Further Analysis}
\begin{figure}[h!]
    \centering
\includegraphics[width=0.49\textwidth]{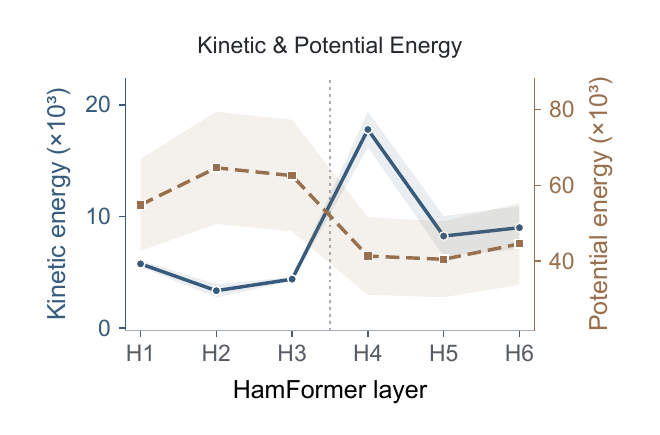}
    \includegraphics[width=0.49\textwidth]{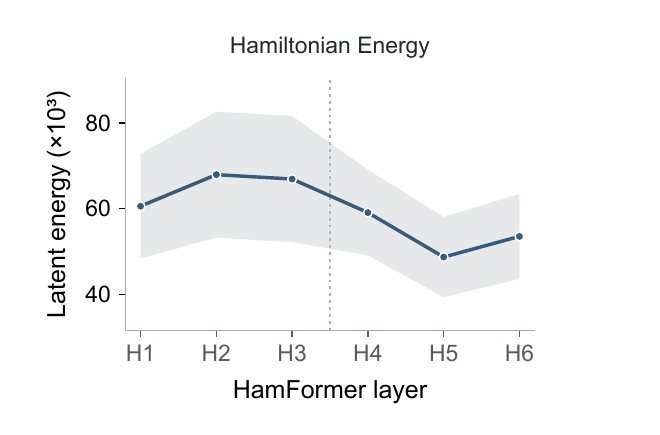}
\caption{
Layer-wise evolution of the latent kinetic energy, potential energy, and total Hamiltonian evaluated on 1,000 random molecules from the Molecule3D test set. Solid curves denote the mean, and shaded regions represent one standard deviation across molecules. The dashed line marks the transition between Stage 1 (layers H1--H3) and Stage 2 (layers H4--H6).}
\label{fig:ablation-energy}
\end{figure}

\textbf{Momentum Evolution over Layers.} 
Figure~\ref{fig:ablation-energy} examines the six HamFormer layers separated into two alternating stages (H1--H3 for Stage 1 and H4--H6 for Stage 2). As expected in a dissipative physical system, the latent dynamics exhibit a clear two-stage relaxation behavior: 
1) across the network as a whole, both the potential energy and total Hamiltonian follow an overall downward trajectory, achieving net energy dissipation toward a stable ground-state equilibrium; 
2) the latent kinetic energy remains relatively suppressed in Stage 1, but experiences a prominent transient surge at the onset of Stage 2 (layer H4, marked by the dashed line) as stored potential energy is rapidly unleashed, followed by smooth dissipation across the remaining layers.

\begin{figure}[htbp]
\centering
\raisebox{-7pt}{\includegraphics[width=0.51\textwidth]{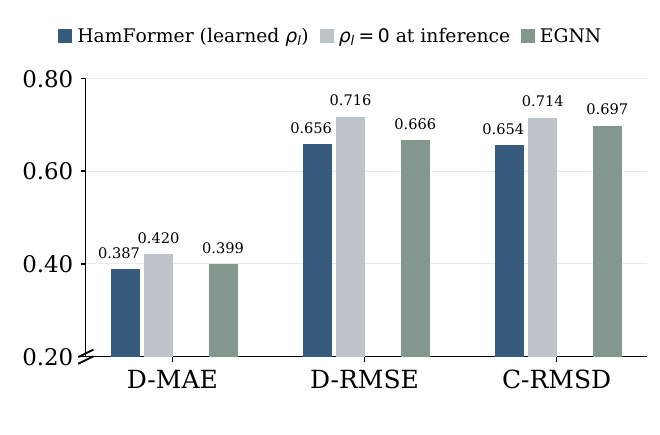}}%
\includegraphics[width=0.47\textwidth]{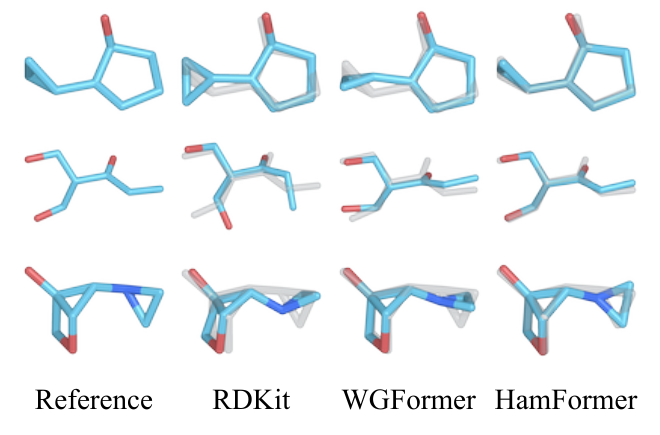}
\caption{
Qualitative conformation visualizations and quantitative ablation studies on Molecule3D.  
\textbf{Left:} Quantitative ablation on the test set comparing the full HamFormer with zeroed inference momentum ($\rho_l = 0$) and the EGNN baseline.
\textbf{Right:} Visual comparison of predicted 3D molecular geometries against DFT references for RDKit, WGFormer, and HamFormer.
}
\label{fig:ablation-Molecule3D}
\vspace{-10pt}
\end{figure}

\textbf{Architectural and Momentum Ablations.}
Figure~\ref{fig:ablation-Molecule3D} (left) systematically investigates the necessity of individual components on the Molecule3D random test set: 
(1) Inference-time momentum retention: Zeroing out the learned momentum at inference ($\rho_l = 0$) induces pronounced performance degradation across all metrics, with test C-RMSD rising from $0.654\,\text{\AA}{}$ to $0.714\,\text{\AA}{}$ and D-MAE increasing from $0.387\,\text{\AA}{}$ to $0.420\,\text{\AA}{}$. This demonstrates that preserving momentum across layers acts as an indispensable dynamical memory, effectively preventing error accumulation during iterative refinement. 
(2) Hamiltonian block ablation: Removing HamFormer and using pure EGNN, which eliminates phase-space feature updates and relies solely on local message-passing—causes notable performance drops (e.g., test C-RMSD rises to $0.697\,\text{\AA}{}$). This confirms that our phase-space Hamiltonian flow provides an essential non-local inductive bias for capturing global molecular geometry.

\textbf{Conformation Visualizations.}
Figure~\ref{fig:ablation-Molecule3D} (right) presents visual comparisons of predicted 3D molecular conformations aligned against ground-truth reference structures for representative molecules. Compared WGFormer, the conformations generated by HamFormer consistently align more closely with the true reference geometries, exhibiting fewer steric clashes and more accurate bond lengths and angles.

\section{Conclusion and Future Work}
We presented HamFormer, which brings Hamiltonian-inspired latent dynamics into SE(3)-equivariant conformation refinement through the joint evolution of features, momentum, and transport-based relations. 
Its integration with EGNN yields improved refinement on QM9 and Molecule3D, suggesting that structured latent dynamics can complement geometric message passing. 
These findings motivate further study
of the balance between representational capacity and computational efficiency, particularly as molecular size and structural complexity increase. Extending this framework beyond small molecules remains a promising direction for assessing its scalability and broader applicability.

\clearpage
\bibliography{references}
\clearpage
\appendix
\startcontents[appendix]

\section*{Appendix Contents}
\begingroup
  \hypersetup{linkcolor=black}
  \printcontents[appendix]{}{1}{\setcounter{tocdepth}{2}}
\endgroup

\clearpage
\section{Theoretical Details}\label{app:derivations}

Section~\ref{app:hamiltonian-dynamics} derives the density equations, their characteristics, and conservation properties. Sections~\ref{app:particle-representation} and~\ref{app:reference-gradients} establish the particle representation of the learned Hamiltonian and compute its force and drift. Section~\ref{app:splitting} derives the Lie--Trotter updates and examines momentum retention, and Section~\ref{app:equivariance} proves SE(3)-equivariance of the complete pipeline. Section~\ref{app:conformation-optimization} presents the molecular refinement algorithms and supervised objective. Finally, Section~\ref{app:training-implementation} details Sinkhorn scaling, rotational alignment, and computational complexity. Continuum calculations assume sufficient regularity to justify differentiation and integration by parts.

\subsection{Hamiltonian Dynamics on Density Space}\label{app:hamiltonian-dynamics}

Consider the Wasserstein Hamiltonian in~(\ref{eq:hf-density-energy}) on a periodic domain $\Omega$,
\begin{equation*}
\mathcal H_W(\rho,S)=\frac12\int_\Omega\|\nabla S(\bm{x})\|^2\rho(\bm{x})\,\mathrm d\bm{x}+\mathcal F(\rho).
\end{equation*}
Let $r$ be a smooth density perturbation with $\int_\Omega r\,\mathrm d\bm{x}=0$, and let $\psi$ be a smooth perturbation of $S$. Writing $D$ for a directional derivative, the two variations are
\begin{align*}
D_\rho\mathcal H_W(\rho,S)[r] &= \int_\Omega\left(\frac12\|\nabla S\|^2+\mathcal F'(\rho)\right)r\,\mathrm d\bm{x},\\ D_S\mathcal H_W(\rho,S)[\psi] &= \int_\Omega\rho\,\nabla S\cdot\nabla\psi\,\mathrm d\bm{x}=-\int_\Omega\nabla\cdot(\rho\nabla S)\,\psi\,\mathrm d\bm{x},
\end{align*}
where periodicity removes the boundary term. Since density perturbations have zero integral, the first variation with respect to $\rho$ is determined up to a spatial constant.

The canonical pairing identifies the density derivative with the variation in $S$, and the potential derivative with the negative variation in $\rho$~\citep{chow2020wasserstein}. Choosing the time-dependent additive constant in $S_t$ gives
\begin{equation*}
\begin{cases}\partial_t\rho_t=-\nabla\cdot(\rho_t\nabla S_t),\\[2pt]\partial_tS_t=-\dfrac12\|\nabla S_t\|^2-\mathcal F'(\rho_t),\end{cases}
\end{equation*}
where the first equation transports probability with velocity $\nabla S_t$, while the second determines the evolution of the conjugate potential.

Let $\bm{x}(t)$ satisfy $\dot{\bm{x}}(t)=\nabla S_t(\bm{x}(t))$, and define $\bm{p}(t)=\nabla S_t(\bm{x}(t))$. The chain rule along this trajectory yields
\begin{align*}
\dot{\bm{p}}(t) &= \nabla\partial_tS_t(\bm{x}(t))+\nabla^2S_t(\bm{x}(t))\dot{\bm{x}}(t)\\ &= \nabla\left(\partial_tS_t+\frac12\|\nabla S_t\|^2\right)(\bm{x}(t))\\ &= -\nabla\mathcal F'(\rho_t)(\bm{x}(t)).
\end{align*}
Thus, with $h_t(\bm{x},\bm{p})=\frac12\|\bm{p}\|^2+\mathcal F'(\rho_t)(\bm{x})$, the characteristic equations are
\begin{equation*}
\dot{\bm{x}}=\nabla_{\bm{p}}h_t=\bm{p},\qquad \dot{\bm{p}}=-\nabla_{\bm{x}}h_t=-\nabla\mathcal F'(\rho_t)(\bm{x}),
\end{equation*}
which recover Hamilton's equations through the classical Hamilton--Jacobi construction~\citep[Sec.~3.3.1]{evans2022partial}. Integrating the continuity equation over $\Omega$ gives
\begin{equation*}
\frac{\mathrm d}{\mathrm dt}\int_\Omega\rho_t\,\mathrm d\bm{x}=-\int_\Omega\nabla\cdot(\rho_t\nabla S_t)\,\mathrm d\bm{x}=0.
\end{equation*}
If $\mathcal F$ has no explicit time dependence, differentiating the Hamiltonian and integrating by parts also gives
\begin{align*}
\frac{\mathrm d}{\mathrm dt}\mathcal H_W(\rho_t,S_t) &= \int_\Omega\left(\frac12\|\nabla S_t\|^2+\mathcal F'(\rho_t)\right)\partial_t\rho_t\,\mathrm d\bm{x}+\int_\Omega\rho_t\nabla S_t\cdot\nabla\partial_tS_t\,\mathrm d\bm{x}\\ &= \int_\Omega\left(-\partial_tS_t\,\partial_t\rho_t+\partial_t\rho_t\,\partial_tS_t\right)\mathrm d\bm{x}=0.
\end{align*}
The conserved quantity is the total functional $\mathcal H_W$. The characteristic Hamiltonian $h_t$ can still depend on time through $\rho_t$.

\subsection{Entropic Interaction and Particle Representation}\label{app:particle-representation}

The learned Hamiltonian in~(\ref{eq:mean-field-hamiltonian}) combines a parametrized kinetic energy density $\kappa_\theta$ with an entropic transport potential. We first verify that this energy is finite and then derive its particle representation.

For the interaction term, the inequality
\begin{equation*}
\left|\bm{z}_h(\bm{x})^\top\bm{z}_h(\bm{x}')\right|\leq\frac12\|\bm{z}_h(\bm{x})\|^2+\frac12\|\bm{z}_h(\bm{x}')\|^2
\end{equation*}
and the finite second moments of the fixed marginals bound the affinity integral uniformly over admissible couplings. The logarithmic relation term is bounded because $t_h$ is bounded and $\varepsilon_{\mathrm{num}}>0$. Relative entropy is nonnegative, and the product coupling $\nu\otimes\nu$ is admissible with zero relative entropy. These observations give finite upper and lower bounds for the interaction value. The kinetic density has at most quadratic growth, so its integral is finite as well.

For distinct fixed atom identifiers $\xi_i$, consider the empirical measures
\begin{equation*}
\mu_N=\frac1N\sum_{i=1}^N\delta_{(\xi_i,\bm{x}_i,\bm{p}_i)}\quad\text{and}\quad\nu_N=\frac1N\sum_{i=1}^N\delta_{(\xi_i,\bm{x}_i)}.
\end{equation*}
The identifiers distinguish atoms with identical features. For a coupling of $\nu_N$ with itself, let $\pi_{ij}/N$ denote the mass assigned to the ordered atom pair $(i,j)$, where
\begin{equation*}
\bm{\pi}\in\bm{\Pi}:=\left\{\bm{\pi}\in\R_+^{N\times N}:\bm{\pi}\mathbf{1}=\bm{\pi}^\top\mathbf{1}=\mathbf{1}\right\},
\end{equation*}
so that $\bm{\pi}$ has unit row and column sums, whereas $\bm{\pi}/N$ is a probability coupling. With $(\bm{T}_h)_{ij}=t_h(\xi_i,\xi_j)$ and $\bm{Z}_h=\bm{X}\bm{W}_h+\mathbf{1}\bm{b}_h^\top$, the sampled affinity is
\begin{equation*}
\bm{S}_h=\frac{\bm{Z}_h\bm{Z}_h^\top}{\sqrt{d_h}}-\log(\bm{T}_h+\varepsilon_{\mathrm{num}}),
\end{equation*}
with finite transport value $\Phi_{\varepsilon_{\mathrm{OT}}}(\bm{S}_h)$ from~(\ref{eq:ot-value}). The following identity accounts for the difference between the distributional and matrix entropy conventions.

\begin{lemma}[\bf Particle normalization]
\label{lem:particle-normalization}
For $\mu_N$ and its corresponding relation tensor,
\begin{equation*}
H_{\theta,N}=N\mathcal H_\theta[\mu_N\mid t]+\varepsilon_{\mathrm{OT}}N(1+\log N)\sum_{h=1}^H\frac{\sqrt{d_h}}2,
\end{equation*}
where the additive correction is independent of $\bm{X}$ and $\bm{P}$.
\end{lemma}

\begin{proof}
The product measure $\nu_N\otimes\nu_N$ assigns mass $1/N^2$ to each ordered atom pair. Using $0\log0=0$ and $\sum_{ij}\pi_{ij}=N$, a coupling with masses $\pi_{ij}/N$ satisfies
\begin{equation*}
\operatorname{KL}(\pi_h\|\nu_N\otimes\nu_N)=\sum_{i,j}\frac{\pi_{ij}}N\log\left(\frac{\pi_{ij}/N}{1/N^2}\right)=\frac1N\sum_{i,j}\pi_{ij}\log\pi_{ij}+\log N.
\end{equation*}
Consequently,
\begin{align*}
N\mathcal V_{\varepsilon_{\mathrm{OT}},h}[\nu_N\mid t_h] &= \max_{\bm{\pi}\in\bm{\Pi}}\Bigl\{\sum_{i,j}\pi_{ij}(S_h)_{ij}-\varepsilon_{\mathrm{OT}}\sum_{i,j}\pi_{ij}\log\pi_{ij}\Bigr\}-\varepsilon_{\mathrm{OT}}N\log N\\ &= \Phi_{\varepsilon_{\mathrm{OT}}}(\bm{S}_h)-\varepsilon_{\mathrm{OT}}N(1+\log N).
\end{align*}
The kinetic term satisfies
\begin{equation*}
N\int\kappa_\theta(\bm{p})\,\mathrm d\mu_N=\sum_{i=1}^N\kappa_\theta(\bm{p}_i)=K_\theta(\bm{P}),
\end{equation*}
and summing the interaction identity over heads with weights $\sqrt{d_h}/2$ proves the result.
\end{proof}

This is an exact evaluation on empirical measures and does not require a limit as $N\to\infty$. After total-energy normalization, the additive constant has no effect on the feature or momentum gradients.

\subsection{Gradients of the Reference Hamiltonian}\label{app:reference-gradients}

For matrices of the same size, write $\langle\bm{A},\bm{B}\rangle=\operatorname{tr}(\bm{A}^\top\bm{B})$. The entropy term makes the finite coupling problem strictly concave in its optimization variable.

\begin{lemma}[\bf Transport gradient]
\label{lem:transport-gradient}
For finite $\bm{S}$ and $\varepsilon_{\mathrm{OT}}>0$, the problem in~(\ref{eq:ot-value}) has a unique positive optimizer $\bm{\pi}^\star$, and
\begin{equation*}
\nabla_{\bm{S}}\Phi_{\varepsilon_{\mathrm{OT}}}(\bm{S})=\bm{\pi}^\star.
\end{equation*}
If $\bm{S}$ is symmetric, then $\bm{\pi}^\star$ is symmetric.
\end{lemma}

\begin{proof}
The admissible set $\bm{\Pi}$ is nonempty and compact, and the objective is continuous under $0\log0=0$. A maximizer therefore exists, and strict concavity of the entropy term gives uniqueness.

If an optimizer has a zero entry, interpolate it with $\mathbf{1}\mathbf{1}^\top/N$. For a small interpolation weight $\tau>0$, the entropy gain at a zero entry has order $-\tau\log\tau$, while the remaining changes have order $\tau$. The objective therefore increases for sufficiently small $\tau$, contradicting optimality.

At a fixed coupling, the derivative of the objective with respect to $\bm{S}$ is $\bm{\pi}$. Danskin's theorem and uniqueness give the stated gradient~\citep{danskin1966theory}. For symmetric $\bm{S}$, transposition preserves feasibility and the objective. Uniqueness then implies $\bm{\pi}^\star=(\bm{\pi}^\star)^\top$.
\end{proof}

Fix $\theta$ and $\bm{\mathsf{T}}$. For head $h$, the score differential is
\begin{equation*}
\mathrm d\bm{S}_h=\frac{(\mathrm d\bm{Z}_h)\bm{Z}_h^\top+\bm{Z}_h(\mathrm d\bm{Z}_h)^\top}{\sqrt{d_h}},
\end{equation*}
where $\mathrm d\bm{Z}_h=(\mathrm d\bm{X})\bm{W}_h$. Applying Lemma~\ref{lem:transport-gradient} gives
\begin{align*}
\mathrm d\left[\frac{\sqrt{d_h}}2\Phi_{\varepsilon_{\mathrm{OT}}}(\bm{S}_h)\right] &= \frac12\left\langle\bm{\pi}_h^\star,(\mathrm d\bm{Z}_h)\bm{Z}_h^\top+\bm{Z}_h(\mathrm d\bm{Z}_h)^\top\right\rangle= \frac12\left\langle\bigl(\bm{\pi}_h^\star+(\bm{\pi}_h^\star)^\top\bigr)\bm{Z}_h,\mathrm d\bm{Z}_h\right\rangle\\ &= \frac12\left\langle\bigl(\bm{\pi}_h^\star+(\bm{\pi}_h^\star)^\top\bigr)\bm{Z}_h\bm{W}_h^\top,\mathrm d\bm{X}\right\rangle.
\end{align*}
Summing over heads yields
\begin{equation*}
\nabla_{\bm{X}}V_\theta(\bm{X}\mid\bm{\mathsf{T}})=\frac12\sum_{h=1}^H\bigl(\bm{\pi}_h^\star+(\bm{\pi}_h^\star)^\top\bigr)\bm{Z}_h\bm{W}_h^\top.
\end{equation*}
The factor $\sqrt{d_h}/2$ cancels the score normalization and averages the contributions from the two feature arguments. If each $\bm{T}_h$ is symmetric, then $\bm{S}_h$ and its optimal plan are symmetric, so the force reduces to
\begin{equation*}
-\nabla_{\bm{X}}V_\theta(\bm{X}\mid\bm{\mathsf{T}})=-\sum_{h=1}^H\bm{\pi}_h^\star\bm{Z}_h\bm{W}_h^\top.
\end{equation*}
These derivatives use the exact optimizer and hold relations fixed.

For the kinetic term, the primitive of leaky-ReLU is
\begin{equation*}
\phi_{0.1}(u)=\int_0^u\sigma_{0.1}(s)\,\mathrm ds=\begin{cases}\frac12u^2,&u\geq0,\\[2pt]\frac{0.1}{2}u^2,&u<0,\end{cases}
\end{equation*}
so $\phi_{0.1}'(u)=\sigma_{0.1}(u)$. Under the row convention for $\bm{P}$,
\begin{equation*}
K_\theta(\bm{P})=\sum_{i,r}\phi_{0.1}\bigl((\bm{P}\bm{B}+\mathbf{1}\bm{b}^\top)_{ir}\bigr).
\end{equation*}
Its differential is
\begin{align*}
\mathrm dK_\theta &= \left\langle\sigma_{0.1}(\bm{P}\bm{B}+\mathbf{1}\bm{b}^\top),(\mathrm d\bm{P})\bm{B}\right\rangle= \left\langle\sigma_{0.1}(\bm{P}\bm{B}+\mathbf{1}\bm{b}^\top)\bm{B}^\top,\mathrm d\bm{P}\right\rangle,
\end{align*}
and hence
\begin{equation*}
\nabla_{\bm{P}}K_\theta(\bm{P})=\sigma_{0.1}(\bm{P}\bm{B}+\mathbf{1}\bm{b}^\top)\bm{B}^\top.
\end{equation*}
The transpose appears because both linear maps arise from differentiating the same scalar energy. This gradient converts momentum into feature velocity. Momentum and velocity coincide for $K_\theta(\bm{P})=\frac12\|\bm{P}\|_F^2$, but need not coincide for the learned energy.

The primitive $\phi_{0.1}$ is continuously differentiable with a Lipschitz derivative, but is only piecewise twice differentiable. Calculations requiring a classical Hessian therefore apply within regions where no kinetic preactivation crosses zero.

\subsection{Lie--Trotter Splitting and the Implemented Layer}\label{app:splitting}

For fixed parameters and relations, write $H_{\theta,N}=K_\theta+V_\theta$. Hamilton's equations are
\begin{equation*}
\begin{cases}\dot{\bm{X}}=\nabla_{\bm{P}}K_\theta(\bm{P}),\\[2pt]\dot{\bm{P}}=-\nabla_{\bm{X}}V_\theta(\bm{X}\mid\bm{\mathsf{T}}).\end{cases}
\end{equation*}
Along a differentiable solution,
\begin{equation*}
\frac{\mathrm d}{\mathrm dt}H_{\theta,N}=\left\langle\nabla_{\bm{X}}V_\theta,\nabla_{\bm{P}}K_\theta\right\rangle-\left\langle\nabla_{\bm{P}}K_\theta,\nabla_{\bm{X}}V_\theta\right\rangle=0,
\end{equation*}
so the autonomous reference flow conserves its Hamiltonian.

The potential subflow changes momentum at fixed features, while the kinetic subflow advances features at fixed momentum. Their exact maps over a step $\eta>0$ are
\begin{align*}
\varphi_V^\eta(\bm{X},\bm{P}) &= \bigl(\bm{X},\bm{P}-\eta\nabla_{\bm{X}}V_\theta(\bm{X}\mid\bm{\mathsf{T}})\bigr),\\ \varphi_K^\eta(\bm{X},\bm{P}) &= \bigl(\bm{X}+\eta\nabla_{\bm{P}}K_\theta(\bm{P}),\bm{P}\bigr).
\end{align*}
Applying $\varphi_V^\eta$ first and $\varphi_K^\eta$ second gives $(\bm{X}^{k+1},\bm{P}^{k+1})=\varphi_K^\eta\circ\varphi_V^\eta(\bm{X}^k,\bm{P}^k)$, which expands to
\begin{equation*}
\begin{cases}\bm{P}^{k+1}=\bm{P}^k-\eta\nabla_{\bm{X}}V_\theta(\bm{X}^k\mid\bm{\mathsf{T}}),\\[2pt]\bm{X}^{k+1}=\bm{X}^k+\eta\nabla_{\bm{P}}K_\theta(\bm{P}^{k+1}).\end{cases}
\end{equation*}
This is the momentum-first symplectic Euler step in~(\ref{eq:hf-symplectic-euler}). For sufficiently smooth energies, Lie--Trotter composition has first-order accuracy, while the symmetric composition $\varphi_V^{\eta/2}\circ\varphi_K^\eta\circ\varphi_V^{\eta/2}$ gives the second-order St\"ormer--Verlet method~\citep{hairer2006geometric}.

Symplecticity follows from the subflow Jacobians. After vectorizing $\bm{X}$ and $\bm{P}$ and suppressing evaluation points, these have the block forms
\begin{equation*}
D\varphi_V^\eta=\begin{pmatrix}\bm{I}&\bm{0}\\-\eta\nabla^2V_\theta&\bm{I}\end{pmatrix},\qquad D\varphi_K^\eta=\begin{pmatrix}\bm{I}&\eta\nabla^2K_\theta\\\bm{0}&\bm{I}\end{pmatrix}.
\end{equation*}
Symmetry of the Hessians gives $(D\varphi)^\top\bm{J}D\varphi=\bm{J}$ for $\bm{J}=\left(\begin{smallmatrix}\bm{0}&\bm{I}\\-\bm{I}&\bm{0}\end{smallmatrix}\right)$, and composition preserves this identity.\footnote{This calculation assumes twice differentiable energies and applies to the learned kinetic map within its smooth activation regions.} The resulting numerical maps need not conserve the original Hamiltonian exactly.

Momentum retention can be introduced through the linear decay subflow $\dot{\bm{P}}=-\gamma\bm{P}$, with $\gamma\geq0$. Its exact solution is $\bm{P}\leftarrow e^{-\gamma\eta}\bm{P}$. Applying decay before the kick gives
\begin{equation*}
\begin{cases}\bm{P}^{k+1}=e^{-\gamma\eta}\bm{P}^k-\eta\nabla_{\bm{X}}V_\theta(\bm{X}^k\mid\bm{\mathsf{T}}),\\[2pt]\bm{X}^{k+1}=\bm{X}^k+\eta\nabla_{\bm{P}}K_\theta(\bm{P}^{k+1}),\end{cases}
\end{equation*}
with $\gamma=0$ recovering the conservative splitting. Applying decay after the kick would also scale the newly added force.

For the corresponding continuous damped system, the energy balance is
\begin{equation*}
\frac{\mathrm d}{\mathrm dt}H_{\theta,N}=-\gamma\left\langle\bm{P},\nabla_{\bm{P}}K_\theta(\bm{P})\right\rangle.
\end{equation*}
For $K_\theta(\bm{P})=\frac12\|\bm{P}\|_F^2$, this becomes $-\gamma\|\bm{P}\|_F^2$, and the feature dynamics satisfy
\begin{equation*}
\ddot{\bm{X}}+\gamma\dot{\bm{X}}+\nabla_{\bm{X}}V_\theta(\bm{X}\mid\bm{\mathsf{T}})=0.
\end{equation*}
For a learned kinetic energy with a bias, the sign of $\langle\bm{P},\nabla_{\bm{P}}K_\theta(\bm{P})\rangle$ is not automatic. Linear momentum decay therefore requires an additional condition to imply energy dissipation.

These calculations concern the reference system with fixed parameters and relations. The learned update in~(\ref{eq:implemented-kick}) follows the same momentum-first order, with the implementation differences described in the main text. {Algorithm~\ref{alg:hamformer-layer} gives the complete layer computation.}

\subsection{SE(3)-Equivariance of the Complete Pipeline}\label{app:equivariance}

The Hamiltonian states $\bm{X}$ and $\bm{P}$ are invariant latent features, while molecular coordinates are updated separately. We hold atom attributes, pair types, and validity masks fixed under rigid motions. During training, we compare transformations using the same feature dropout masks.

\begin{lemma}[\bf SE(3)-equivariance]
\label{lem:equivariance}
The complete coordinate predictor $f$, including centering and centroid restoration, satisfies
\begin{equation*}
f(\bm{C}\bm{R}^\top+\mathbf{1}\bm{t}^\top)=f(\bm{C})\bm{R}^\top+\mathbf{1}\bm{t}^\top
\end{equation*}
on valid atoms for every $\bm{R}\in\mathrm{SO}(3)$ and $\bm{t}\in\R^3$.
\end{lemma}

\begin{proof}
Under $\bm{c}_i'=\bm{R}\bm{c}_i+\bm{t}$, pairwise distances satisfy
\begin{equation*}
\|\bm{c}_i'-\bm{c}_j'\|=\|\bm{R}(\bm{c}_i-\bm{c}_j)\|=\|\bm{c}_i-\bm{c}_j\|.
\end{equation*}
Relation initialization uses only pair types and these distances, so the initialized relations are unchanged. Atom embeddings and zero initial momentum are unchanged as well. Every HamFormer operation acts on invariant quantities, so induction over its layers gives identical features, momentum, and relations for the original and transformed inputs. This argument holds for any fixed number of Sinkhorn iterations.

At EGNN layer $q$, the current and stage-entry distances remain invariant. The atom features and relation vectors are also invariant, so the pair messages, updated features, and scalar coordinate weights $w_{ij}^{(q)}$ are unchanged. Writing $\bm{c}_i^{\prime(q)}=\bm{R}\bm{c}_i^{(q)}+\bm{t}$, the coordinate update gives
\begin{align*}
\bm{c}_i^{\prime(q+1)} &= \bm{R}\bm{c}_i^{(q)}+\bm{t}+\sum_j w_{ij}^{(q)}\frac{\bm{R}(\bm{c}_i^{(q)}-\bm{c}_j^{(q)})}{\|\bm{c}_i^{(q)}-\bm{c}_j^{(q)}\|+\varepsilon_{\mathrm{dist}}}\\ &= \bm{R}\bm{c}_i^{(q+1)}+\bm{t}.
\end{align*}
Thus the coordinate update is equivariant, as in the EGNN construction~\citep{satorras2021n}. Its feature output remains invariant and can enter the next HamFormer stage. Passing momentum through EGNN unchanged preserves its invariance.

For mask $\bm{m}$ and $N=\mathbf{1}^\top\bm{m}$, the input centroid is $\bm{c}_{\mathrm{ctr}}=(\bm{C}^{\mathrm{in}})^\top\bm{m}/N$. It transforms as $\bm{c}_{\mathrm{ctr}}'=\bm{R}\bm{c}_{\mathrm{ctr}}+\bm{t}$ and $\bm{C}_{\mathrm c}'=\bm{C}_{\mathrm c}\bm{R}^\top$.
The centered network output therefore rotates by $\bm{R}$. Restoring the input centroid gives $\widehat{\bm{C}}'=\widehat{\bm{C}}\bm{R}^\top+\mathbf{1}\bm{t}^\top$, which proves the claim.
\end{proof}

Predicted pairwise distances are consequently invariant. This argument depends on the transformation properties of the network operations and does not require an exact Hamiltonian integrator or exact transport marginals.

\subsection{HamFormer for Molecular Conformation Optimization}\label{app:conformation-optimization}

Given atom types $\bm{a}$, pair types $\bm{E}=(e_{ij})$, and input coordinates $\bm{C}^{\mathrm{in}}$, the atom embedding serves as the encoder and the final EGNN as the SE(3)-equivariant coordinate decoder. The network operates on centered coordinates and restores the input centroid to obtain $\widehat{\bm{C}}$, with pairwise distances $\widehat{\bm{D}}$. Algorithms~\ref{alg:hamformer-layer} and~\ref{alg:hamformer} specify the layer update and pipeline.

Both algorithms update states in place. For a valid-atom mask $\bm{m}\in\{0,1\}^{N_{\mathrm{pad}}}$, let $N=\mathbf{1}^\top\bm{m}$. Masked Sinkhorn targets marginals $\bm{m}/N$, and geometric statistics use only valid atoms. {Let $L_E$ denote the number of HamFormer--EGNN stage pairs. For $l=1,\ldots,L_E$, the stage maps $\operatorname{HamFormer}_l$ and $\operatorname{EGNN}_l$ apply their layers in order. Each EGNN stage records its incoming coordinates to compute the fixed stage-entry distances.} The map $\operatorname{InitRelation}$ implements~(\ref{eq:initial-relation}).

\begin{figure}[htbp]
\algrenewcommand{\algorithmicrequire}{\ensuremath{\triangleright}\ \textbf{Require:}}
\algrenewcommand{\algorithmicreturn}{\ensuremath{\triangleright}\ \textbf{return}}
\centering
\begin{minipage}[t]{0.51\textwidth}
\vspace{0pt}
\resizebox{\linewidth}{!}{%
\begin{minipage}{1.11\linewidth}
\begin{algorithm}[H]
\caption{HamFormer layer}
\label{alg:hamformer-layer}
\footnotesize
\begin{algorithmic}[1]
\Require $\bm{X},\bm{P},\bm{\mathsf{T}},\bm{m}$, parameters $\theta_\ell$, iterations $I$
\State $N\gets\mathbf{1}^\top\bm{m}$
\For{$h=1,\ldots,H$}
\State $\bm{Z}_h\gets\LN_{\mathrm{in},\ell}(\bm{X})\bm{W}_h^{(\ell)}+\mathbf{1}(\bm{b}_h^{(\ell)})^\top$
\State $\bm{S}\gets\bm{Z}_h\bm{Z}_h^\top/\sqrt{d_h}-\log(\bm{T}_h+\varepsilon_{\mathrm{num}})$
\State $\bm{T}_h\gets N\,\Sink_{\varepsilon_{\mathrm{OT}},I}(\bm{S},\bm{m})$
\EndFor
\State $\bm{P}\gets-\tanh(\alpha_\ell)\bm{P}+\LN_{P,\ell}\bigl(-\textstyle\sum_{h=1}^H\bm{T}_h\bm{Z}_h(\bm{W}_h^{(\ell)})^\top\bigr)$
\State $\bm{X}\gets\bm{X}+\LN_{X,\ell}\bigl[\sigma_{0.1}(\bm{P}\bm{B}_\ell+\mathbf{1}\bm{b}_\ell^\top)\bm{B}_\ell^\top\bigr]$
\State $\bm{X}\gets\bm{X}+\operatorname{FFN}_\ell(\bm{X})$
\State \Return $\bm{X},\bm{P},\bm{\mathsf{T}}$
\end{algorithmic}
\end{algorithm}
\end{minipage}%
}
\end{minipage}
\hfill
\begin{minipage}[t]{0.48\textwidth}
\vspace{0pt}
\resizebox{\linewidth}{!}{%
\begin{minipage}{1.12\linewidth}
\begin{algorithm}[H]
\caption{Conformation refinement}
\label{alg:hamformer}
\footnotesize
\begin{algorithmic}[1]
\Require $\bm{a},\bm{E},\bm{C}^{\mathrm{in}},\bm{m}$
\State $N\gets\mathbf{1}^\top\bm{m},\quad \bm{X}\gets\operatorname{Embed}(\bm{a}),\quad \bm{P}\gets\bm{0}$
\State $\bm{\mathsf{T}}\gets\operatorname{InitRelation}(\bm{E},\bm{C}^{\mathrm{in}},\bm{m})$
\State $\bm{c}_{\mathrm{ctr}}\gets(\bm{C}^{\mathrm{in}})^\top\bm{m}/N$
\State $\bm{C}\gets\bm{C}^{\mathrm{in}}-\mathbf{1}\bm{c}_{\mathrm{ctr}}^\top$
\For{$l=1,\ldots,L_E$}
\State $(\bm{X},\bm{P},\bm{\mathsf{T}})\gets\operatorname{HamFormer}_l(\bm{X},\bm{P},\bm{\mathsf{T}},\bm{m})$
\State $(\bm{X},\bm{C})\gets\operatorname{EGNN}_l(\bm{X},\bm{C},\bm{\mathsf{T}},\bm{m})$
\EndFor
\State $\widehat{\bm{C}}\gets\bm{C}+\mathbf{1}\bm{c}_{\mathrm{ctr}}^\top\;$ and $\;\widehat{\bm{D}}\gets[\|\widehat{\bm{c}}_i-\widehat{\bm{c}}_j\|]_{ij}$
\State \Return $\widehat{\bm{C}},\widehat{\bm{D}}$
\end{algorithmic}
\end{algorithm}
\end{minipage}%
}
\end{minipage}
\end{figure}

For ground-state conformation prediction~\citep{wang2025wgformer}, we minimize distance and aligned coordinate errors relative to target distances $\bm{D}^{\mathrm{gt}}$ and coordinates $\bm{C}^{\mathrm{gt}}$:
\begin{equation}
\mathcal L_{\mathrm{train}}(\theta)=\frac{1}{N^2}\sum\nolimits_{(i,j)\in\mathcal I_{\mathrm{pair}}}\bigl|\widehat D_{ij}-D_{ij}^{\mathrm{gt}}\bigr|+\frac{\lambda}{N}\min\nolimits_{\bm{R}\in\mathrm{SO}(3)}\bigl\|\widehat{\bm{C}}_{\mathrm c}-\bm{C}_{\mathrm c}^{\mathrm{gt}}\bm{R}^\top\bigr\|_F^2,
\label{eq:training-loss}
\end{equation}
where the subscript $\mathrm c$ denotes centering over the $N$ valid atoms, $\mathcal I_{\mathrm{pair}}$ contains all ordered valid atom pairs, and $\lambda\geq0$ weights the coordinate loss. The first term measures absolute distance errors, while the second measures squared coordinate errors after optimal rotation. During training, this rotation is computed from detached predictions and held fixed during backpropagation. {Section}~\ref{app:training-implementation} details rotational alignment and masked Sinkhorn scaling. All network parameters are trained jointly, with gradients propagated through the relation updates, normalization, and Sinkhorn iterations.

\subsection{Transport Computation, Training, and Complexity}\label{app:training-implementation}

For the positive optimizer of~(\ref{eq:ot-value}), introduce row and column multipliers ${\zeta_i}$ and $\beta_j$. Stationarity gives
\begin{equation*}
S_{ij}-\varepsilon_{\mathrm{OT}}\log\pi_{ij}-{\zeta_i}-\beta_j=0,
\end{equation*}
so the optimizer has the form $\pi_{ij}=u_i\exp(S_{ij}/\varepsilon_{\mathrm{OT}})v_j$, where $u_i=e^{-{\zeta_i}/\varepsilon_{\mathrm{OT}}}$ and $v_j=e^{-\beta_j/\varepsilon_{\mathrm{OT}}}$. Sinkhorn scaling alternately adjusts $\bm{u}$ and $\bm{v}$ to satisfy the row and column constraints~\citep{cuturi2013sinkhorn,peyre2019computational}.

The implementation targets probability marginals $\bm{m}/N$. Write $\bm{\pi}^{\mathrm{prob}}$ for the resulting matrix, with invalid rows and columns excluded from normalization. On the valid block, the scaling updates are
\begin{equation*}
u_i\leftarrow\frac{1/N}{\sum_j\exp(S_{ij}/\varepsilon_{\mathrm{OT}})v_j},\qquad v_j\leftarrow\frac{1/N}{\sum_i u_i\exp(S_{ij}/\varepsilon_{\mathrm{OT}})}.
\end{equation*}
The relation update returns $N\bm{\pi}^{\mathrm{prob}}$, {targeting unit row and column sums on valid atoms}. With finitely many iterations, both sets of constraints need not hold exactly. Alternating row and column normalization also need not preserve symmetry at intermediate iterations.

For the relation-conditioned score {in~(\ref{eq:param-hamiltonian})},
\begin{equation*}
\exp(\bm{S}_h/\varepsilon_{\mathrm{OT}})=\exp\left(\frac{\bm{Z}_h\bm{Z}_h^\top}{\varepsilon_{\mathrm{OT}}\sqrt{d_h}}\right)\odot(\bm{T}_h+\varepsilon_{\mathrm{num}})^{-1/\varepsilon_{\mathrm{OT}}},
\end{equation*}
where $\odot$ and the power act entrywise. At $\varepsilon_{\mathrm{OT}}=1$, {the exponentiated feature affinity is divided entrywise by $\bm{T}_h+\varepsilon_{\mathrm{num}}$ before normalization}.

{For the coordinate term in~(\ref{eq:training-loss}), both coordinate matrices are restricted to valid atoms and centered before alignment.} Expanding the coordinate error gives
\begin{align*}
\big\|\widehat{\bm{C}}_{\mathrm c}-\bm{C}_{\mathrm c}^{\mathrm{gt}}\bm{R}^\top\big\|_F^2 = \|\widehat{\bm{C}}_{\mathrm c}\|_F^2+\|\bm{C}_{\mathrm c}^{\mathrm{gt}}\|_F^2-2\operatorname{tr}\big(\bm{R}^\top\widehat{\bm{C}}_{\mathrm c}^\top\bm{C}_{\mathrm c}^{\mathrm{gt}}\big).
\end{align*}
The first two terms are independent of $\bm{R}$. Let $\widehat{\bm{C}}_{\mathrm c}^\top\bm{C}_{\mathrm c}^{\mathrm{gt}}=\bm{U}_{\mathrm{svd}}\bm{\Sigma}\bm{V}_{\mathrm{svd}}^\top$ be a singular value decomposition with singular values in decreasing order. A minimizing proper rotation is
\begin{equation*}
\bm{R}^\star=\bm{U}_{\mathrm{svd}}\operatorname{diag}\left(1,1,\det(\bm{U}_{\mathrm{svd}}\bm{V}_{\mathrm{svd}}^\top)\right)\bm{V}_{\mathrm{svd}}^\top,
\end{equation*}
where the determinant correction enforces $\bm{R}^\star\in\mathrm{SO}(3)$. At points where the minimized coordinate loss is differentiable, for example when the minimizing rotation is unique, the envelope identity gives
\begin{align*}
&\nabla_{\widehat{\bm{C}}_{\mathrm c}}\Big[{\frac{\lambda}{N}}\min_{\bm{R}\in\mathrm{SO}(3)}\big\|\widehat{\bm{C}}_{\mathrm c}-\bm{C}_{\mathrm c}^{\mathrm{gt}}\bm{R}^\top\big\|_F^2\Big]={\frac{2\lambda}{N}}\Big(\widehat{\bm{C}}_{\mathrm c}-\bm{C}_{\mathrm c}^{\mathrm{gt}}(\bm{R}^\star)^\top\Big).
\end{align*}
Computing the exact optimal rotation from detached predictions and treating it as fixed therefore yields the gradient of the minimized value at such points. Degenerate configurations can have multiple minimizing rotations, so differentiability is not asserted everywhere.

\begin{remark}[\bf Computational complexity]\label{rem:complexity}
Consider one dense HamFormer layer with $N$ valid atoms, feature width $d$, head widths $d_h$, FFN width $d_{\mathrm{FF}}$, and $I$ Sinkhorn iterations. Head projections cost $O(Nd\sum_h d_h)$. Forming scores and aggregating transport messages cost $O(N^2\sum_h d_h)$, and Sinkhorn scaling costs $O(IHN^2)$. The kinetic map and FFN cost $O(Nd^2)$ and $O(Nd\,d_{\mathrm{FF}})$, respectively. The total time complexity is therefore $O\bigl(Nd\sum_h d_h+N^2\sum_h d_h+IHN^2+Nd^2+Nd\,d_{\mathrm{FF}}\bigr)$, which reduces to $O\bigl(Nd^2+N^2d+IHN^2+Nd\,d_{\mathrm{FF}}\bigr)$ when $\sum_h d_h=d$.

The relation tensor requires $O(HN^2)$ storage. Including features, projected heads, and FFN activations gives forward activation storage $O\bigl(HN^2+Nd+N\sum_h d_h+Nd_{\mathrm{FF}}\bigr)$. Training requires additional saved intermediates depending on the differentiation and recomputation strategy. For dense padded tensors, replace $N$ by $N_{\mathrm{pad}}$ unless padded entries are excluded from computation. EGNN cost scales with the number of evaluated atom pairs at fixed MLP widths, and the complete network cost is obtained by summing over its layers.
\end{remark}

\section{Experimental Details}
\subsection{Datasets}
We use the QM9 splits and the Molecule3D random split datasets, following the benchmark protocol used by WGFormer.
The processed QM9 validation and test sets contain 9,900 and 10,697 molecules, respectively. The corresponding Molecule3D sets contain 779,918 and 779,915 molecules. 

\subsection{Evaluation}
Following WGFormer, we use the following metrics to evaluate the quality of the predicted molecular structures. 
\begin{itemize}
\item \textbf{D-MAE} and \textbf{D-RMSE}:
      The mean absolute error and root mean square error between
      the predicted pairwise distances $\widehat{D}_{ij}$ and
      ground-truth distances $D_{ij}^{\mathrm{gt}}$, both computed
      from the corresponding Cartesian coordinates:
       \[
        \begin{aligned}
        \mathrm{D\text{-}MAE}
        &=
        \frac{1}{N^2}
        \sum\nolimits_{(i,j)\in\mathcal I_{\mathrm{pair}}}
        \bigl|\widehat D_{ij}-D_{ij}^{\mathrm{gt}}\bigr|,\\
        \mathrm{D\text{-}RMSE}
        &=
        \sqrt{
            \frac{1}{N^2}
            \sum\nolimits_{(i,j)\in\mathcal I_{\mathrm{pair}}}
            \bigl(\widehat D_{ij}-D_{ij}^{\mathrm{gt}}\bigr)^2
        }.
        \end{aligned}
        \]
\item \textbf{C-RMSD}:
            The coordinate root mean square deviation between the predicted coordinates $\widehat{\bm C}$ and ground-truth coordinates $\bm C^{\mathrm{gt}}$ after optimal rigid alignment:
            \[
            \mathrm{C\text{-}RMSD}
            =
            \sqrt{
                \frac{1}{N}
                \min_{\bm R\in\mathrm{SO}(3),\, \bm t\in\mathbb{R}^3}
                \left\|
                    \widehat{\bm C}
                    -
                    (\bm C^{\mathrm{gt}}\bm R^\top + \bm 1_N \bm t^\top)
                \right\|_F^2
            }.
            \]
            For this metric, hydrogen atoms are removed, $N$ counts the
        retained atoms, and $\bm C_{\mathrm c}^{\mathrm{gt}}$ is
        understood to use the symmetry-equivalent atom ordering that
        minimizes the aligned RMSD.     
\end{itemize}
\clearpage
\subsection{Hyperparameters}
Table~\ref{tab:app-training-setting} summarizes the full-model training settings used on the QM9 and Molecule3D random datasets, including the peak learning rate, scheduled epochs, warmup fraction, evaluation checkpoint epoch, number of training GPUs, batch size per GPU, global batch size, gradient accumulation steps, and random seed. 
\begin{table}[htbp]
\centering\small
\caption{Full-model training settings.}
\label{tab:app-training-setting}
\begin{tabular}{lcc}
\toprule
Setting & QM9 & Molecule3D random\\
\midrule
Peak learning rate & $8\times10^{-4}$ & $4\times10^{-4}$\\
Scheduled epochs & 500 & 40\\
Warmup fraction & 0.06 & 0.05\\
Evaluated checkpoint epoch & 294 & 40\\
Training GPUs & 8 & 2\\
Batch size per GPU & 32 & 64\\
Global batch size & 256 & 128\\
Gradient accumulation & 1 & 1\\
Random seed & 1 & 1\\
\bottomrule
\end{tabular}
\end{table}
\subsection{Ablation Study on Model Components}
To further validate the effectiveness of our proposed design, we perform module ablation studies on the validation and test sets of both QM9 and Molecule3D random split. As shown in Table~\ref{tab:abl-module-detail}, removing the HamFormer feature block and the $P\&T$ branch leads to a prominent performance degradation across all metrics. 
This significant drop clearly demonstrates that the introduced components are essential for maintaining the model's high representation capacity and prediction accuracy.

\begin{table}[htbp]
\centering
\caption{Module ablation on the test sets. ``Feature block'' denotes the
module applied before each EGNN stage.}
\label{tab:abl-module-detail}
\small
\setlength{\tabcolsep}{3pt}
\begin{tabular}{lccrrrrrr}
\toprule
& Feature & $P \& T$ & \multicolumn{3}{c}{QM9} & \multicolumn{3}{c}{Molecule3D random}\\
\cmidrule(lr){4-6}\cmidrule(lr){7-9}
Variant & block & branch & D-MAE & D-RMSE & C-RMSD & D-MAE & D-RMSE & C-RMSD\\
\midrule
Full model & HamFormer & $\checkmark$ 
&\textbf{0.218} 
& \textbf{0.419} 
& \textbf{0.198} 
& \textbf{0.387}
& \textbf{0.656}
& \textbf{0.654}
\\
EGNN-only
& $\times$ 
& $\times$ 
& 0.233 
& 0.427 
& 0.239 
& 0.399
& 0.666 
& 0.697
\\
\bottomrule
\end{tabular}
\end{table}

\subsection{Inference-Time Momentum-Retention Ablation}
To further validate the importance of the inference-time momentum-retention mechanism, we investigate the effect of the momentum factor on the Molecule3D (random split) dataset. As shown in Table~\ref{tab:appendix-retention-ablation}, setting the retention factor to zero ($\rho_l = 0$) results in a substantial performance drop across both validation and test sets compared to using the learned momentum parameter ($\text{Learned } \rho_l$). This prominent degradation strongly confirms the crucial role and importance of the momentum-retention mechanism during inference.

\begin{table}[htbp]
\centering
\caption{Inference-time momentum-retention ablation on Molecule3D random.}
\label{tab:appendix-retention-ablation}
\small
\setlength{\tabcolsep}{4pt}
\begin{tabular}{lrrrrrr}
\toprule
& \multicolumn{3}{c}{Validation} & \multicolumn{3}{c}{Test}\\
\cmidrule(lr){2-4}\cmidrule(lr){5-7}
Inference setting & D-MAE & D-RMSE & C-RMSD & D-MAE & D-RMSE & C-RMSD\\
\midrule
Learned $\rho_l$ 
& \textbf{0.386}
& \textbf{0.653}
& \textbf{0.652}
& \textbf{0.387}
& \textbf{0.656}
& \textbf{0.654}
\\
$\rho_l=0$ 
& 0.418
& 0.684
& 0.712
& 0.420
& 0.716
& 0.714
\\
\bottomrule
\end{tabular}
\end{table}

\subsection{Layer-wise Energy Evolution on the QM9 Dataset}
\begin{figure}[t]
    \centering
    \includegraphics[width=0.49\textwidth]{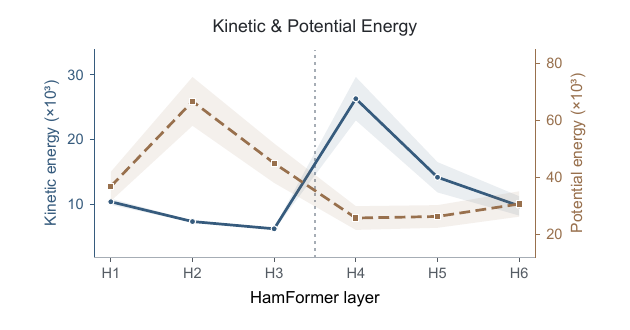}
    \includegraphics[width=0.49\textwidth]{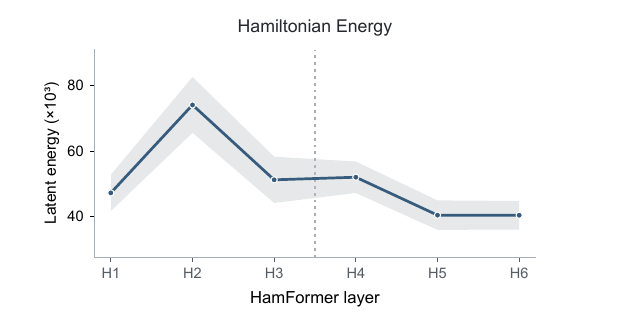}
\caption{Layer-wise evolution of the latent kinetic energy, potential energy, and total Hamiltonian evaluated on 1,000 randomly selected molecules from the QM9 test set.}
\label{fig:ablation-energy-qm9}
\end{figure}

Figure~\ref{fig:ablation-energy-qm9} presents the layer-wise evolution of the latent kinetic energy, potential energy, and total Hamiltonian on the QM9 test set. The energy profiles exhibit clear changes across HamFormer layers, particularly around the transition from Stage 1 to Stage 2. This observation is consistent with the corresponding results on Molecule3D and further suggests that the Hamiltonian dynamics are progressively transformed throughout the network.

\subsection{Additional Visualizations}
\begin{figure}[h]
    \centering
\includegraphics[width=0.65\linewidth]{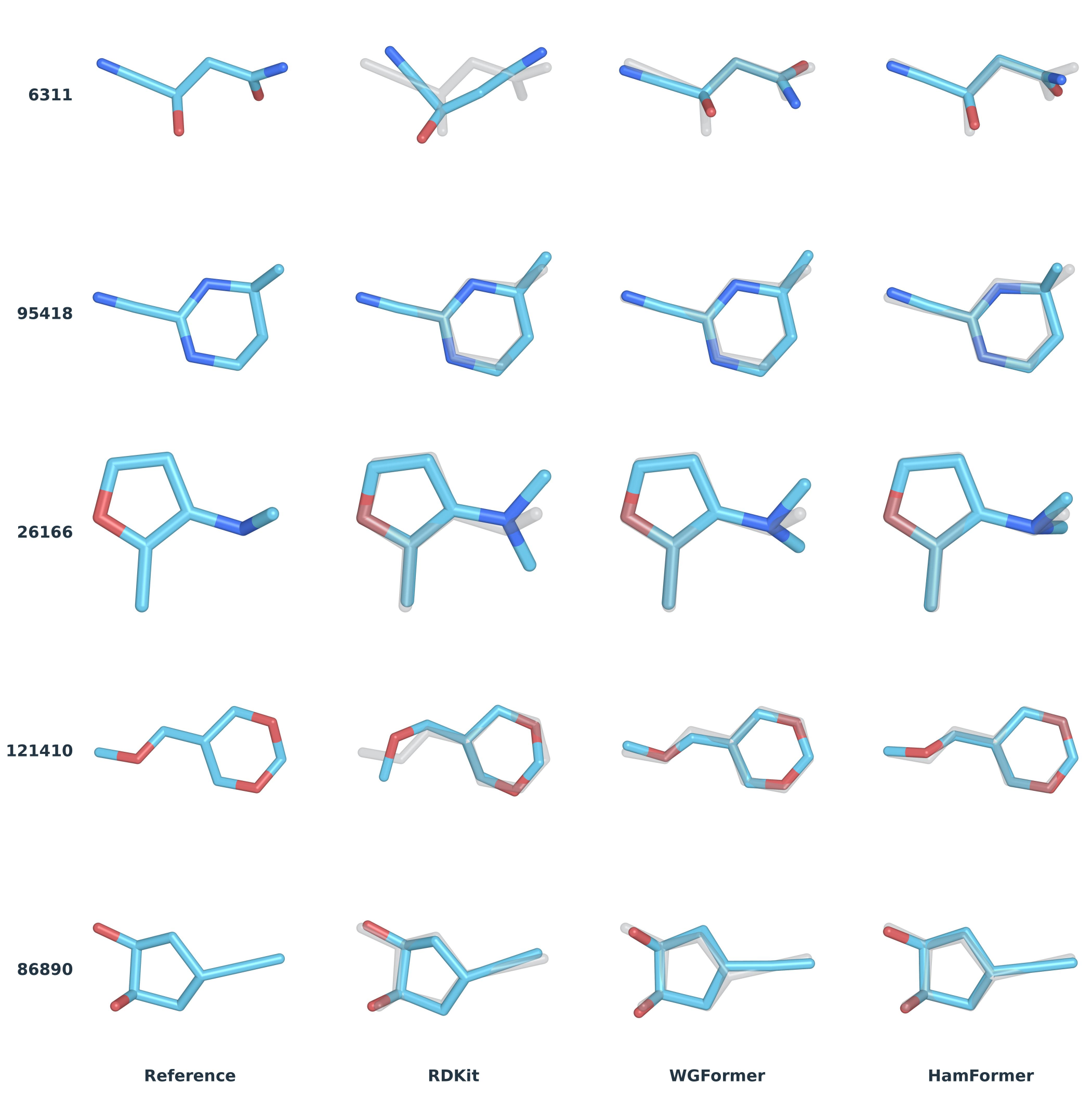}
    \caption{The visualizations for the ground-state conformations predicted by RDKit, WGFormer, and HamFormer.}
    \label{fig:placeholder}
\end{figure}
\end{document}